\pdfoutput=1

\documentclass[11pt]{article}

\usepackage[final]{acl}

\usepackage{times}
\usepackage{latexsym}

\usepackage[T1]{fontenc}

\usepackage[utf8]{inputenc}

\usepackage{microtype}

\usepackage{inconsolata}

\usepackage{graphicx}

 \usepackage{xspace}
\usepackage{amsmath}  
\usepackage{amssymb}  
\usepackage{amsfonts} 
\usepackage{enumitem}
\usepackage{booktabs} 
\usepackage{geometry} 
\usepackage{multirow} 
\usepackage{titletoc}
\usepackage{float}
\usepackage[skins]{tcolorbox}
\usepackage{fontawesome}

\usepackage[table]{xcolor}
\definecolor{niceblue}{RGB}{62, 138, 204}

\newcommand\ours{ThinkARM\xspace}

\title{Schoenfeld's Anatomy of Mathematical Reasoning by Language Models}

\author{%
    \textbf{Ming Li*}, 
    \textbf{Chenrui Fan*}, 
    \textbf{Yize Cheng*},
    \textbf{Soheil Feizi},
    \textbf{Tianyi Zhou}\\
    {University of Maryland, College Park}\\
    \texttt{\{minglii,cfan42,yzcheng,sfeizi\}@umd.edu, tianyi.david.zhou@gmail.com} \\
    \faGithub~Project: \url{https://github.com/MingLiiii/ThinkARM}
}

\begin{document}
\maketitle
\renewcommand{\thefootnote}{}
\footnotetext{*Co-First Authors. }
\renewcommand{\thefootnote}{\arabic{footnote}}

\begin{abstract}
  Large language models increasingly expose reasoning traces, yet their underlying cognitive structure and steps remain difficult to identify and analyze beyond surface-level statistics. We adopt Schoenfeld's Episode Theory as an inductive, intermediate-scale lens and introduce \textbf{\ours} (\textit{\textbf{A}natomy of \textbf{R}easoning in \textbf{M}odels}), a scalable framework that \textbf{explicitly abstracts reasoning traces into functional reasoning steps} such as \textit{Analysis}, \textit{Explore}, \textit{Implement}, \textit{Verify}, etc. When applied to mathematical problem solving by diverse models, this abstraction reveals reproducible thinking dynamics and structural differences between reasoning and non-reasoning models, which are not apparent from token-level views. We further present two diagnostic case studies showing that exploration functions as a critical branching step associated with correctness, and that efficiency-oriented methods selectively suppress evaluative feedback steps rather than uniformly shortening responses. Together, our results demonstrate that episode-level representations make reasoning steps explicit, enabling systematic analysis of how reasoning is structured, stabilized, and altered in modern language models.
\end{abstract}

\section{Introduction}

Large language models (LLMs) have demonstrated remarkable performance on complex reasoning tasks~\cite{openai2024o1, marjanovic2025deepseekr1, comanici2025gemini25pushingfrontier, qwen3}, particularly when generating explicit chains of thought (CoT)~\cite{wei2023chainofthoughtpromptingelicitsreasoning}. Consequently, evaluation of reasoning models has predominantly focused on outcome-oriented metrics such as accuracy, solution length, or aggregate token counts~\cite{ lightman2023letsverifystepstep,jiang2025mmecotbenchmarkingchainofthoughtlarge}. While these measures are effective for comparing final performance, they provide limited insight into how these models organize their reasoning traces and how different reasoning behaviors emerge across models.

It is still unclear which parts of a generated chain-of-thought correspond to problem understanding, exploration, execution, or verification, making it difficult to interpret model behavior beyond surface statistics. This opacity is particularly salient when studying ``overthinking''~\cite{chen2025think23overthinkingo1like, fan2025missing, kumar2025overthinkslowdownattacksreasoning}, where longer or more elaborate reasoning does not necessarily translate into improved correctness~\cite{feng2025characterizeseffectivereasoningrevisiting}, yet the underlying thinking dynamics and structure are hard to characterize systematically. Various reasoning behaviors that are often discussed \textit{conceptually}, e.g., abstract planning versus concrete execution, open-ended exploration versus evaluative checking, lack rigorous formulation and quantitative comparison across models.

\begin{figure*}[t]
    \centering
    \includegraphics[width=0.94\linewidth]{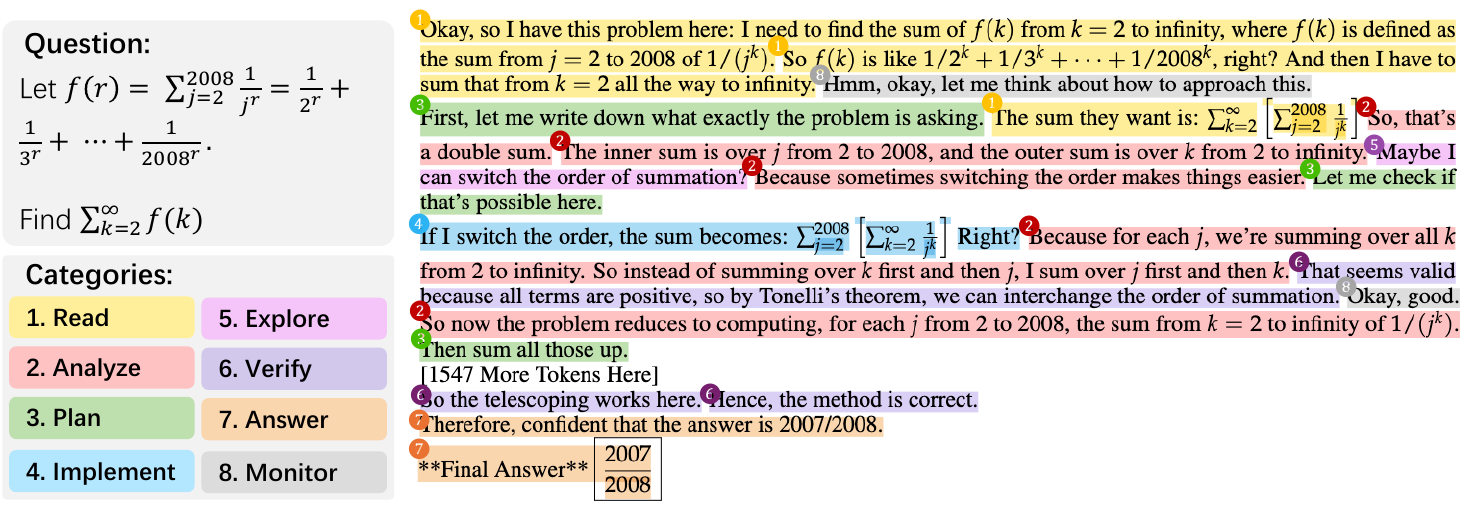}
    \vspace{-2.2mm}
    \caption{A condensed example of a reasoning trace that is annotated in our framework. Each sentence in response is tagged with one of the eight episode categories.
    }
    \label{fig:intro}
    \vspace{-4.2mm}
\end{figure*}

To better interpret such reasoning traces, a natural question is whether they contain meaningful structure at an intermediate level of abstraction beyond individual tokens. Motivated by prior work~\cite{li-etal-2025-understanding} that introduced Schoenfeld's Episode Theory~\cite{schoenfeld2014mathematical} as a framework for characterizing problem-solving behaviors, we adopt episode-level representations as an \textit{\textbf{inductive lens}} for analyzing LLM reasoning traces. Episode Theory conceptualizes problem solving in terms of functional episodes, thereby providing an interpretable intermediate-scale representation that bridges low-level token statistics and high-level reasoning intents. A condensed example is shown in Figure \ref{fig:intro}.

While earlier work~\cite{li-etal-2025-understanding} first leveraged this theory to reasoning models, their scope is limited to one model and one dataset. Thus, it remains unclear whether such an episode-level abstraction can reveal systematic, reproducible structure in LLM reasoning at scale. In this work, we build upon this foundation and extend episode-level annotation to a broader, comparative setting, using it to examine what principal patterns emerge when diverse reasoning traces are viewed through a theory-grounded abstraction.
Our study is organized around three research questions.
\textbf{RQ1:} \textit{Do reasoning traces exhibit consistent linguistic and dynamic structure when viewed through episode-level representations?}
\textbf{RQ2:} \textit{How do reasoning dynamics differ across models and reasoning styles, as indicated by episode sequencing and transition patterns?}
\textbf{RQ3:} \textit{Can we utilize this framework to analyze the correlation of thinking patterns to final correctness, and the structural differences between overthinking and efficient models? }

To address these questions, we apply \textbf{\ours} (\textit{\textbf{A}natomy of \textbf{R}easoning in \textbf{M}odels}), an episode-level annotation framework grounded in Schoenfeld's Episode Theory~\cite{harskamp2007schoenfeld}, to a large-scale analysis of mathematical reasoning traces from a diverse set of LLMs. Concretely, we curate a corpus of $410,991$ sentences generated by $15$ models solving $100$ problems from a subset of Omni-MATH~\cite{gao2024omni}. To support reliable automated analysis, we construct a human-verified gold set of $7,067$ sentences and evaluate multiple state-of-the-art LLMs as automatic annotators, and finally select GPT-5~\cite{openai2025gpt5} for full-scale episode labeling due to its strongest agreement with human annotations. This pipeline enables consistent, sentence-level episode annotation at scale and sets up a controlled setting for comparing reasoning dynamics across models and methods.

\paragraph{Contributions.}
We extend a cognitive science-inspired episode annotation framework to an automatic, scalable, sentence-level representation that supports large-scale analysis of reasoning traces and conduct a systematic study of reasoning dynamics across a diverse set of LLMs. Moreover, we demonstrate the practical utility of episode-level representations through downstream case studies on correctness and efficiency, illustrating how reasoning dynamics can be analyzed beyond outcome-based metrics.

\paragraph{Key Findings.} 
\begin{enumerate} [leftmargin=4mm]
    \item When reasoning traces are analyzed at the episode level, \textit{\textbf{a functional progression from abstract reasoning to concrete execution, and finally to evaluative control, consistently emerges.}} Episodes associated with analysis and exploration use more abstract, conceptual language and \textit{\textbf{decrease}} steadily as reasoning progresses, while execution-oriented episodes \textit{\textbf{dominate}} the middle of the trace through sustained concrete operations. In contrast, verification-related episodes are characterized by evaluative and meta-level language and \textit{\textbf{increase}} toward the end of the reasoning process.
    \vspace{-2.2mm}
    \item Comparing reasoning and non-reasoning models, the difference is not merely how many tokens they generate, but how reasoning is structured. \textit{\textbf{Non-reasoning models allocate most of their response trace to execution}}, with episode transitions largely following a feed-forward pattern toward implementation. In contrast, \textit{\textbf{reasoning models distribute effort across analysis, exploration, execution, and verification, and exhibit frequent iterative Explore-Monitor/Verify loops.}} \looseness-1
    \vspace{-2.2mm}
    \item Through our correctness-oriented case study, we find that \textit{\textbf{exploration reflects uncertainty and serves as a critical branching point}}: correct solutions more often route exploration into monitoring or re-analysis, whereas incorrect solutions tend to continue execution or terminate prematurely after exploration. 
    \vspace{-2.2mm}
    \item Through our efficiency-oriented case study, we find that \textit{\textbf{different efficient reasoning methods selectively suppress evaluation-oriented episodes and feedback loops, leading to varying degrees of divergence}} from the reasoning patterns of the base model. Episode-level analysis thus reveals which episodes can be removed to gain efficiency, beyond token-level pruning.
\end{enumerate}

\noindent
Together, these findings make explicit a range of intermediate-scale reasoning behaviors \textit{\textbf{that are often discussed intuitively but rarely characterized structurally.}}

\section{The \ours Framework}

In this section, we formalize our analytical methodology. We first ground our approach in cognitive science theory in mathematical problem solving and then detail the implementation of \ours, a scalable, automated pipeline to quantify the reasoning dynamics of LLMs with high precision.

\subsection{Schoenfeld's Episode Theory}

To scientifically dissect the reasoning traces of LLMs, we ground our framework in Schoenfeld's Episode Theory~\cite{schoenfeld2014mathematical}, a seminal framework in cognitive science originally developed to decode the black box of human mathematical problem-solving. 
Schoenfeld's theory is descriptive, derived from the rigorous analysis of hundreds of hours of videotaped ``think-aloud'' protocols. By observing students and mathematicians tackling mathematical problems, they identified that the performance is not distinguished by superior domain knowledge alone, but by the dynamic regulation of that knowledge. The theory frames problem-solving as a temporally ordered sequence of ``episodes'' that reveal the solver's evolving goal structure and metacognitive decisions\footnote{More discussion on mathematical problem solving in cognitive science can be found in Appendix \ref{sec:related_work}}. 

Schoenfeld's framework ultimately consists of 7 episodes: the original 6 episodes, including \textit{Read}, \textit{Analyze}, \textit{Plan}, \textit{Implement}, \textit{Explore}, and \textit{Verify}, plus a later-added \textit{Monitor} episode that specifically captures the solver's metacognitive behaviors.
Episode theory has since become a foundational analytic lens in mathematics-education research and in studies of human reasoning, offering a fine-grained vocabulary for tracing cognitive control and strategy shifts \citep{harskamp2007schoenfeld}.
\citet{li-etal-2025-understanding} first applied this theory to annotate reasoning traces of LLMs with thinking behaviors.
However, their work is limited to single-task scenarios and single-model case studies and does not provide a systematic analysis of the reasoning dynamics of diverse LLMs. 

\begin{table}[t]
  \centering
  \resizebox{\linewidth}{!}{
    \begin{tabular}{lccccccc}
      \toprule
      & \multicolumn{2}{c}{\textbf{Reasoning}} & \multicolumn{2}{c}{\textbf{Non-Reasoning}} & \multicolumn{2}{c}{\textbf{Overall}} \\
      \cmidrule(lr){2-3} \cmidrule(lr){4-5} \cmidrule(lr){6-7}
      \textbf{Model} & Acc & Kappa & Acc & Kappa & Acc & Kappa \\
      \midrule
      GPT-4.1 & 85.75 & 82.39 & 89.34 & \textbf{85.36} & 86.10 & 82.74 \\
      GPT-5 & \textbf{86.02} & \textbf{82.54} & \textbf{89.34} & 85.35 & \textbf{86.33} & \textbf{82.85} \\
      Gemini-2.5-Flash & 82.45 & 78.21 & 87.16 & 82.35 & 82.90 & 78.67 \\
      Gemini-2.5-Pro & 80.53 & 75.60 & 84.43 & 78.62 & 80.89 & 75.96 \\
      \bottomrule
    \end{tabular}
    }
    \vspace{-2.2mm}
    \caption{
      Performance of candidate annotators on the human-annotated gold set.
      GPT-5 shows the highest agreement with human annotations overall and is used for further large-scale annotation.
    }
    \label{tab:annotation_acc}
    \vspace{-4.2mm}
\end{table}

\subsection{A Refined Taxonomy}

While Schoenfeld's original taxonomy captures key functional stages of human problem solving, it does not include an explicit state for structural convergence. In the context of LLM reasoning, where models are trained to produce a final answer in a prescribed format, this distinction becomes practically important. We therefore extend the taxonomy by introducing an \textit{Answer} episode, resulting in \textit{\textbf{Eight}} episodes, which allows us to explicitly identify when the model commits to producing a final solution and to analyze convergence behavior separately from verification or monitoring, as shown in Figure \ref{fig:intro}.

Prior work \cite{li-etal-2025-understanding} explored episode-based annotation using hierarchical schemes that combine paragraph-level and sentence-level labels. In this study, we adopt sentence-level annotation only, as it provides a uniform granularity that facilitates large-scale aggregation, transition analysis, and comparison across models. This design choice reflects a trade-off toward scalability and analytical simplicity, rather than a claim about the relative merits of different annotation granularities.

\begin{figure*}[t]
    \centering 
    \includegraphics[width=0.98\textwidth]{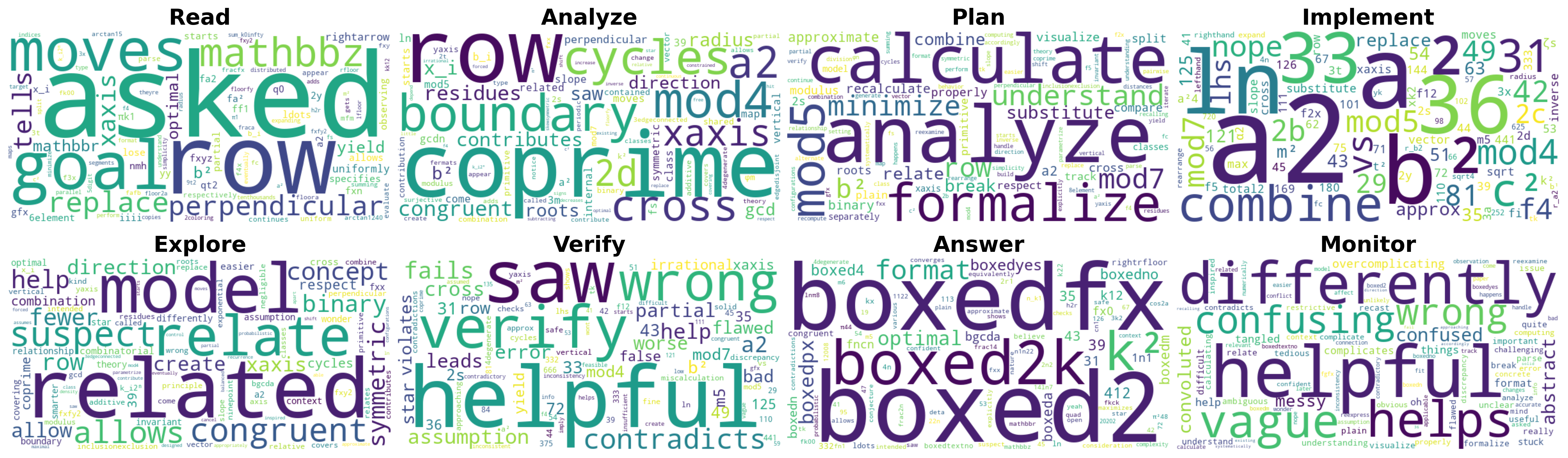} 
    \vspace{-2.2mm}
    \caption{Word clouds visualizing the most frequent semantic tokens for each cognitive episode. \textbf{The distinct lexical distributions highlight the semantically separable cognitive patterns captured by \ours.}
    }
    \vspace{-4.2mm}
    \label{fig:word_cloud} 
    \end{figure*}

\subsection{Data Collection and Gold Annotation}

To construct a diverse and representative dataset for our analysis, we sample problems from Omni-MATH using its domain annotations. Specifically, we stratify the full benchmark by domain and select problems from each group proportionally, aiming to preserve domain coverage while maintaining diversity in problem types and difficulty. This procedure yields a subset of $100$ problems spanning a broad range of mathematical topics.

We then collect reasoning traces utilizing $15$ widely used LLMs\footnote{The complete list of models is in Appendix \ref{sec:implementation_details}} to solve each problem, resulting in $1,500$ responses and a total of $410,991$ sentences. 
The collected traces include both thinking and answering tokens for native reasoning models and their distilled variants, and only answering tokens for standard instruction-following baselines and proprietary reasoning models without accessible thinking tokens. This diverse model coverage enables comparative analysis across model families and reasoning paradigms.

To support the reliable evaluation of automated episode annotation, we construct a human-verified gold set. From the $100$ sampled problems, we select $9$ representative problems following the same domain-based grouping strategy and manually annotate\footnote{The detailed annotation guidebook is in Appendix \ref{sec:refined_annotation_guidebook}} all resulting reasoning traces using the refined episode taxonomy and guidebook. This process produces $7,067$ annotated sentences, which we use as the gold standard for testing the effectiveness of the automatic annotators and selecting the annotation model used in our large-scale analysis.

\subsection{The \ours Framework}

In \ours, we utilize LLMs for automatic annotation.
During the automatic annotation process, we provide the detailed annotation guidebook, which consists of the definition and examples of each episode, along with the previous contexts and annotated categories. 
Furthermore, when annotating each sentence, we not only ask models to generate the annotated category, but also generate the justifications for each annotation to ensure the reliability of the annotation\footnote{The detailed \ours Framework is in Appendix \ref{sec:detailed_ours_framework} and the annotation prompt is in Appendix \ref{sec:annotation_prompt}}. 

For the annotation model selection, we evaluate the performance of several state-of-the-art models, including GPT-4.1~\cite{openai2025gpt4.1}, GPT-5~\cite{openai2025gpt5}, Gemini-2.5-Flash~\cite{comanici2025gemini25pushingfrontier}, and Gemini-2.5-Pro, on the gold standard annotated traces.
The performance of the annotation models is shown in Table \ref{tab:annotation_acc}, containing the accuracy and kappa scores on the traces from reasoning and non-reasoning models. 
Based on the performance, we select GPT-5 for the full-scale annotation and utilize its results for further analysis.

\section{Episode Patterns of Reasoning Models}

With \ours, we now move beyond surface-level performance metrics to empirically analyze the cognitive dynamics of current LLMs.

\begin{figure*}[t]
    \centering 
    \includegraphics[width=0.98\textwidth]{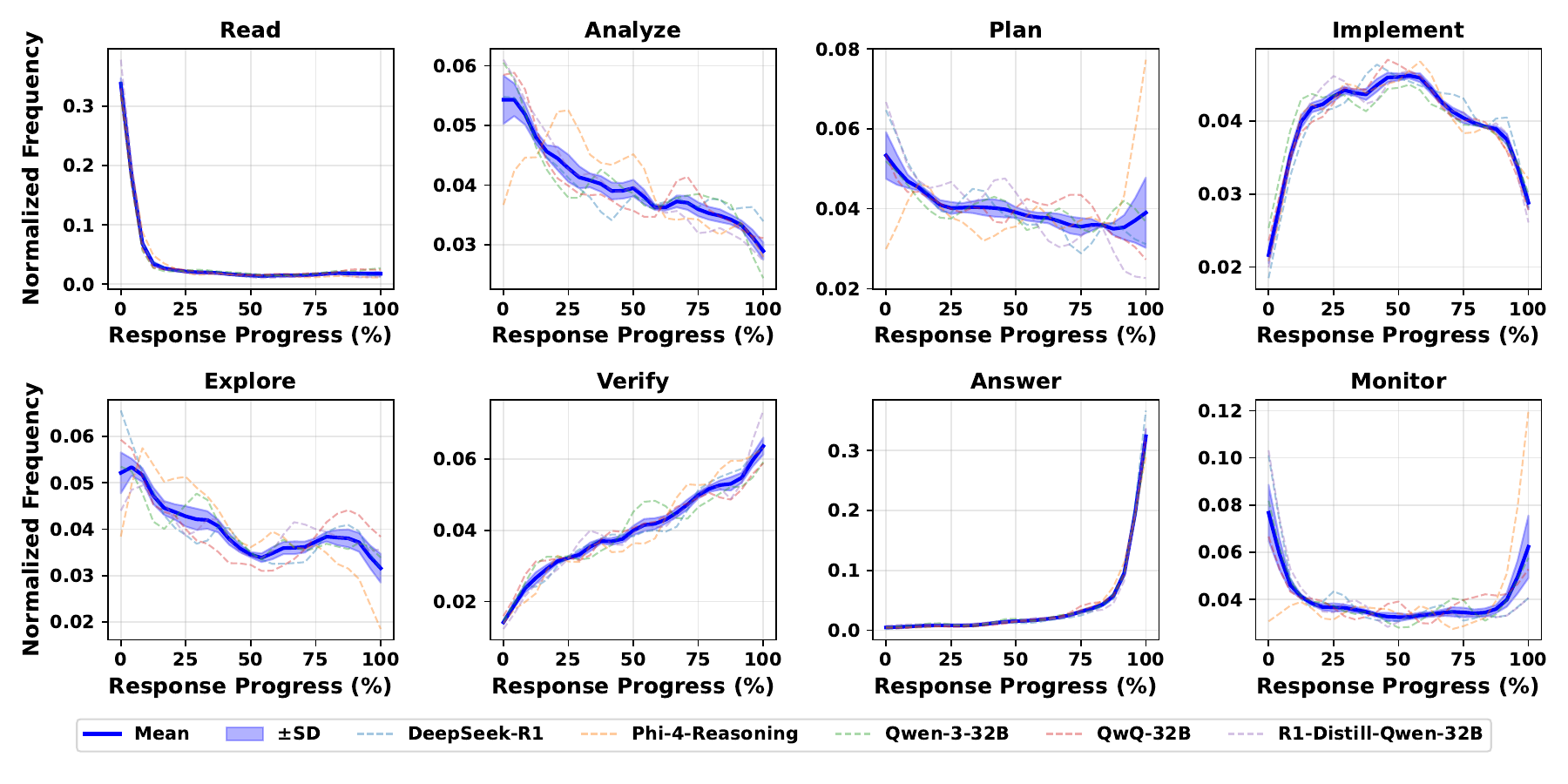} 
    \vspace{-2.2mm}
    \caption{
        Thinking dynamics of cognitive episodes \textbf{reveal a three-phase ``heartbeat'' pattern of reasoning models:} (1) Initialization, dominated by \textit{Read}, \textit{Analyze}, and \textit{Plan}; (2) Execution, where \textit{Implement} peaks; and (3) Convergence, characterized by a surge in \textit{Verify} and \textit{Monitor} before the final \textit{Answer}. 
    } 
    \label{fig:temporal_plot} 
    \vspace{-4.2mm}
    \end{figure*}

\subsection{Episode Divergence}

Our analysis asks whether different functional modes of reasoning, often discussed intuitively, manifest as separable patterns in language. 
Figure~\ref{fig:word_cloud} visualizes the most frequent tokens associated with each episode and shows that episodes occupy distinct lexical regions, suggesting that \textbf{\ours captures meaningful behavioral differences rather than superficial variation.}

\textbf{\textit{Analyze} vs. \textit{Implement}:}
A clear separation emerges between abstract reasoning and concrete execution. Language associated with \textit{Analyze} emphasizes conceptual and structural elements of the problem (e.g., ``coprime'', ``boundary''), reflecting the construction and manipulation of an abstract problem representation. In contrast, \textit{Implement} is dominated by procedural and symbol-level language (e.g., variable names and concrete values), indicating step-by-step execution of a chosen approach. Thus, \ours captures a genuine shift from conceptual thinking to solid implementation in reasoning behavior.

\textbf{\textit{Verify} vs. \textit{Monitor}: }
Two qualitatively different forms of reflection also emerge. \textit{Verify} is characterized by decisive evaluative language (e.g., ``wrong'', ``helpful''), indicating explicit checking of logical correctness or validity. \textit{Monitor}, by contrast, is associated with meta-level expressions of uncertainty or progress tracking (e.g., ``confusing'', ``messy''), reflecting awareness of the state of the reasoning process rather than evaluation of a specific step. These two behaviors that are separated lexically support treating verification and monitoring as distinct reasoning modes, which are important in later temporal and structural analyses.

\textbf{\textit{Plan} vs. \textit{Explore}:}
Forward-looking reasoning also exhibits two separable modes. \textit{Plan} is dominated by directive verbs and goal-oriented language (e.g., ``calculate'', ``formalize''), signaling commitment to a concrete strategy. In contrast, \textit{Explore} is marked by tentative and hypothesis-driven expressions (e.g., ``suspect'', ``maybe''), reflecting open-ended search over possible solution paths. This separation highlights exploration as a distinct behavioral mode characterized by uncertainty, rather than merely an early stage of execution.

\begin{table*}[htb]
    \centering
    \resizebox{\textwidth}{!}{%
    \begin{tabular}{l | cccccccc | cccccccc | c}
    \toprule
    \textbf{Model} & \multicolumn{8}{c|}{\textbf{Percentage (\%)}} & \multicolumn{8}{c|}{\textbf{Token Count (\#)}} & \textbf{Avg} \\
    \cmidrule(lr){2-9} \cmidrule(lr){10-17}
     & \textbf{Read} & \textbf{Ana} & \textbf{Plan} & \textbf{Imp} & \textbf{Exp} & \textbf{Ver} & \textbf{Ans} & \textbf{Mon} & \textbf{Read} & \textbf{Ana} & \textbf{Plan} & \textbf{Imp} & \textbf{Exp} & \textbf{Ver} & \textbf{Ans} & \textbf{Mon} & \textbf{Tok} \\
    \midrule
    
    DeepSeek-R1 & \cellcolor{niceblue!35}3.47 & \cellcolor{niceblue!82}30.75 & \cellcolor{niceblue!45}5.67 & \cellcolor{niceblue!86}36.35 & \cellcolor{niceblue!55}9.21 & \cellcolor{niceblue!57}10.16 & \cellcolor{niceblue!32}2.86 & \cellcolor{niceblue!22}1.54 & \cellcolor{niceblue!68}321 & \cellcolor{niceblue!94}2844 & \cellcolor{niceblue!74}524 & \cellcolor{niceblue!96}3363 & \cellcolor{niceblue!80}852 & \cellcolor{niceblue!81}940 & \cellcolor{niceblue!66}265 & \cellcolor{niceblue!58}142 & 9250 \\
    R1-Distill-Qwen-1.5B & \cellcolor{niceblue!39}4.18 & \cellcolor{niceblue!79}26.93 & \cellcolor{niceblue!39}4.20 & \cellcolor{niceblue!84}34.39 & \cellcolor{niceblue!63}13.23 & \cellcolor{niceblue!60}11.43 & \cellcolor{niceblue!37}3.80 & \cellcolor{niceblue!25}1.84 & \cellcolor{niceblue!71}414 & \cellcolor{niceblue!93}2666 & \cellcolor{niceblue!71}416 & \cellcolor{niceblue!96}3404 & \cellcolor{niceblue!85}1309 & \cellcolor{niceblue!83}1131 & \cellcolor{niceblue!69}376 & \cellcolor{niceblue!62}182 & 9897 \\
    R1-Distill-Qwen-7B & \cellcolor{niceblue!37}3.80 & \cellcolor{niceblue!77}25.82 & \cellcolor{niceblue!40}4.44 & \cellcolor{niceblue!86}36.47 & \cellcolor{niceblue!61}12.24 & \cellcolor{niceblue!61}11.97 & \cellcolor{niceblue!35}3.40 & \cellcolor{niceblue!25}1.86 & \cellcolor{niceblue!68}336 & \cellcolor{niceblue!91}2280 & \cellcolor{niceblue!71}392 & \cellcolor{niceblue!95}3220 & \cellcolor{niceblue!83}1081 & \cellcolor{niceblue!82}1057 & \cellcolor{niceblue!67}300 & \cellcolor{niceblue!60}165 & 8831 \\
    R1-Distill-Qwen-32B & \cellcolor{niceblue!34}3.12 & \cellcolor{niceblue!77}25.23 & \cellcolor{niceblue!41}4.75 & \cellcolor{niceblue!86}36.49 & \cellcolor{niceblue!59}10.94 & \cellcolor{niceblue!58}10.67 & \cellcolor{niceblue!49}6.85 & \cellcolor{niceblue!25}1.95 & \cellcolor{niceblue!67}300 & \cellcolor{niceblue!92}2422 & \cellcolor{niceblue!72}456 & \cellcolor{niceblue!96}3502 & \cellcolor{niceblue!82}1050 & \cellcolor{niceblue!82}1025 & \cellcolor{niceblue!77}658 & \cellcolor{niceblue!62}187 & 9598 \\
    QwQ-32B & \cellcolor{niceblue!33}3.03 & \cellcolor{niceblue!76}24.53 & \cellcolor{niceblue!47}6.28 & \cellcolor{niceblue!85}35.71 & \cellcolor{niceblue!64}13.98 & \cellcolor{niceblue!59}11.05 & \cellcolor{niceblue!34}3.32 & \cellcolor{niceblue!26}2.09 & \cellcolor{niceblue!70}378 & \cellcolor{niceblue!95}3068 & \cellcolor{niceblue!79}785 & \cellcolor{niceblue!99}4466 & \cellcolor{niceblue!88}1748 & \cellcolor{niceblue!85}1382 & \cellcolor{niceblue!71}416 & \cellcolor{niceblue!66}261 & 12504 \\
    Phi-4-Reasoning & \cellcolor{niceblue!39}4.16 & \cellcolor{niceblue!79}27.83 & \cellcolor{niceblue!48}6.42 & \cellcolor{niceblue!86}37.51 & \cellcolor{niceblue!60}11.56 & \cellcolor{niceblue!55}8.91 & \cellcolor{niceblue!31}2.75 & \cellcolor{niceblue!15}0.87 & \cellcolor{niceblue!71}415 & \cellcolor{niceblue!94}2771 & \cellcolor{niceblue!76}639 & \cellcolor{niceblue!97}3734 & \cellcolor{niceblue!83}1151 & \cellcolor{niceblue!80}887 & \cellcolor{niceblue!66}273 & \cellcolor{niceblue!53}86 & 9957 \\
    Qwen-3-32B (R) & \cellcolor{niceblue!32}2.88 & \cellcolor{niceblue!77}25.58 & \cellcolor{niceblue!51}7.79 & \cellcolor{niceblue!86}36.86 & \cellcolor{niceblue!60}11.56 & \cellcolor{niceblue!58}10.54 & \cellcolor{niceblue!33}3.04 & \cellcolor{niceblue!24}1.76 & \cellcolor{niceblue!69}372 & \cellcolor{niceblue!96}3308 & \cellcolor{niceblue!82}1007 & \cellcolor{niceblue!100}4767 & \cellcolor{niceblue!86}1495 & \cellcolor{niceblue!85}1363 & \cellcolor{niceblue!70}393 & \cellcolor{niceblue!64}227 & 12931 \\
    \midrule
    GPT-4o & \cellcolor{niceblue!53}8.49 & \cellcolor{niceblue!74}22.66 & \cellcolor{niceblue!63}12.96 & \cellcolor{niceblue!89}40.77 & \cellcolor{niceblue!30}2.58 & \cellcolor{niceblue!41}4.84 & \cellcolor{niceblue!49}6.94 & \cellcolor{niceblue!13}0.77 & \cellcolor{niceblue!48}59 & \cellcolor{niceblue!60}156 & \cellcolor{niceblue!53}89 & \cellcolor{niceblue!67}281 & \cellcolor{niceblue!34}18 & \cellcolor{niceblue!42}33 & \cellcolor{niceblue!46}48 & \cellcolor{niceblue!22}5 & 690 \\
    Gemini-2.0-Flash & \cellcolor{niceblue!45}5.63 & \cellcolor{niceblue!71}19.24 & \cellcolor{niceblue!40}4.33 & \cellcolor{niceblue!98}61.17 & \cellcolor{niceblue!16}0.97 & \cellcolor{niceblue!40}4.45 & \cellcolor{niceblue!39}4.14 & \cellcolor{niceblue!2}0.08 & \cellcolor{niceblue!49}63 & \cellcolor{niceblue!63}215 & \cellcolor{niceblue!46}48 & \cellcolor{niceblue!77}682 & \cellcolor{niceblue!29}11 & \cellcolor{niceblue!46}50 & \cellcolor{niceblue!46}46 & \cellcolor{niceblue!7}1 & 1115 \\
    Qwen-2.5-32B & \cellcolor{niceblue!45}5.85 & \cellcolor{niceblue!74}22.78 & \cellcolor{niceblue!64}13.58 & \cellcolor{niceblue!91}45.62 & \cellcolor{niceblue!15}0.88 & \cellcolor{niceblue!35}3.39 & \cellcolor{niceblue!50}7.25 & \cellcolor{niceblue!12}0.65 & \cellcolor{niceblue!45}41 & \cellcolor{niceblue!60}160 & \cellcolor{niceblue!54}95 & \cellcolor{niceblue!69}320 & \cellcolor{niceblue!16}6 & \cellcolor{niceblue!38}24 & \cellcolor{niceblue!47}51 & \cellcolor{niceblue!18}5 & 700 \\
    Phi-4 & \cellcolor{niceblue!42}4.87 & \cellcolor{niceblue!66}15.66 & \cellcolor{niceblue!48}6.66 & \cellcolor{niceblue!99}64.12 & \cellcolor{niceblue!4}0.21 & \cellcolor{niceblue!39}4.27 & \cellcolor{niceblue!38}3.89 & \cellcolor{niceblue!6}0.31 & \cellcolor{niceblue!49}64 & \cellcolor{niceblue!63}206 & \cellcolor{niceblue!53}88 & \cellcolor{niceblue!80}844 & \cellcolor{niceblue!15}3 & \cellcolor{niceblue!48}56 & \cellcolor{niceblue!47}51 & \cellcolor{niceblue!19}4 & 1316 \\
    Qwen-3-32B       & \cellcolor{niceblue!49}6.87 & \cellcolor{niceblue!63}13.03 & \cellcolor{niceblue!49}6.82 & \cellcolor{niceblue!100}65.76 & \cellcolor{niceblue!16}0.97 & \cellcolor{niceblue!35}3.42 & \cellcolor{niceblue!31}2.82 & \cellcolor{niceblue!6}0.30 & \cellcolor{niceblue!61}182 & \cellcolor{niceblue!69}345 & \cellcolor{niceblue!61}181 & \cellcolor{niceblue!88}1742 & \cellcolor{niceblue!39}26 & \cellcolor{niceblue!53}91 & \cellcolor{niceblue!50}75 & \cellcolor{niceblue!26}8 & 2649 \\
    \midrule
    Gemini-2.5-Flash & \cellcolor{niceblue!30}2.64 & \cellcolor{niceblue!79}28.04 & \cellcolor{niceblue!49}6.47 & \cellcolor{niceblue!94}52.28 & \cellcolor{niceblue!8}0.42 & \cellcolor{niceblue!49}6.75 & \cellcolor{niceblue!33}3.05 & \cellcolor{niceblue!7}0.35 & \cellcolor{niceblue!49}60 & \cellcolor{niceblue!76}642 & \cellcolor{niceblue!59}148 & \cellcolor{niceblue!83}1197 & \cellcolor{niceblue!28}10 & \cellcolor{niceblue!59}155 & \cellcolor{niceblue!50}70 & \cellcolor{niceblue!26}8 & 2290 \\
    GPT-o1-mini & \cellcolor{niceblue!56}9.42 & \cellcolor{niceblue!75}22.97 & \cellcolor{niceblue!54}8.64 & \cellcolor{niceblue!89}42.01 & \cellcolor{niceblue!13}0.71 & \cellcolor{niceblue!39}4.26 & \cellcolor{niceblue!59}11.32 & \cellcolor{niceblue!12}0.67 & \cellcolor{niceblue!47}53 & \cellcolor{niceblue!57}129 & \cellcolor{niceblue!46}49 & \cellcolor{niceblue!65}237 & \cellcolor{niceblue!19}4 & \cellcolor{niceblue!38}24 & \cellcolor{niceblue!49}64 & \cellcolor{niceblue!18}4 & 563 \\
    GPT-o3-mini & \cellcolor{niceblue!46}6.17 & \cellcolor{niceblue!83}31.87 & \cellcolor{niceblue!52}8.04 & \cellcolor{niceblue!85}35.34 & \cellcolor{niceblue!16}0.97 & \cellcolor{niceblue!41}4.66 & \cellcolor{niceblue!57}9.90 & \cellcolor{niceblue!33}3.04 & \cellcolor{niceblue!49}63 & \cellcolor{niceblue!68}328 & \cellcolor{niceblue!52}83 & \cellcolor{niceblue!70}363 & \cellcolor{niceblue!28}10 & \cellcolor{niceblue!46}48 & \cellcolor{niceblue!55}102 & \cellcolor{niceblue!41}31 & 1027 \\
    \bottomrule
    \end{tabular}%
    }
    \vspace{-2.2mm}
    \caption{
        Episode-level allocation across model families.
        \textbf{Standard instruction-following models are strongly \textit{Implement}-heavy, whereas reasoning models exhibit a more balanced allocation with substantial mass on \textit{Analyze}, \textit{Explore}, and \textit{Verify}. Distilled models largely preserve the teacher's allocation profile across scales.}
    }
    \vspace{-4.2mm}
    \label{tab:epi_ratio}
\end{table*}

\subsection{Temporal Dynamics}

By normalizing each reasoning trace to a 0-100\% progress scale\footnote{Details in Appendix \ref{sec:implementation_details}}, we analyze how episode frequencies evolve over the generation (Figure~\ref{fig:temporal_plot}). A consistent coarse-grained temporal organization emerges, captured by \ours, across reasoning models. We refer to this pattern as the \textit{cognitive heartbeat} of machine reasoning. 
\looseness-1

\textit{\textbf{Early-stage scaffolding:}}
Episodes associated with initial scaffolding (\textit{Read}, \textit{Analyze}, \textit{Plan}, \textit{Explore}) exhibit distinct decay patterns. \textit{Read} drops sharply after the beginning, while \textit{Analyze} and \textit{Plan} decay more gradually, indicating that structural reasoning persists beyond the first few steps rather than being confined to a fixed planning prefix. \textit{Explore} is typically front-loaded as well, consistent with early hypothesis search that narrows as execution proceeds.

\textit{\textbf{Mid-stage execution:}}
\textit{Implement} exhibits a characteristic bell shape trajectory, peaking in the middle of the trace. This pattern suggests that concrete execution occupies the longest continuous region of the reasoning process, providing the backbone onto which other behaviors (e.g., exploration and verification) attach. The model shifts from high-level strategizing to mechanical execution (symbolic manipulation and calculation) as the core engine of the response. 

\textit{\textbf{Late-stage convergence:}}
\textit{Verify} increases steadily over progress, and \textit{Monitor} follows a U-shaped profile with elevated mass near the beginning and end. Together, these trends indicate that evaluative and process-regulatory behaviors become more prominent as generation approaches completion, rather than appearing only as a final end check. Finally, \textit{Answer} remains near-zero for most of the trace and rises sharply near the end, reflecting that answer commitment is concentrated in the terminal stage.

\subsection{Episode Coverage}

Table~\ref{tab:epi_ratio} reports the episode-level allocation of generated tokens. We group models into: (i) open-source reasoning models where full reasoning traces are available, (ii) standard instruction-following models evaluated on their direct responses, and (iii) proprietary reasoning models where only the final responses are observable.

\textbf{\textit{From Doing to Thinking}:}
A clear allocation gap emerges between reasoning and non-reasoning models. Standard instruction-following models allocate the majority of tokens to \textit{Implement}, with minimal mass assigned to \textit{Explore}, \textit{Verify}, or \textit{Monitor}. In contrast, reasoning models exhibit a more balanced profile, allocating substantially more budget to \textit{Analyze} and \textit{Explore}, and maintaining non-trivial allocation to \textit{Verify}. \textbf{This suggests that the distinguishing factor is not merely response length, but how the generation budget is distributed across reasoning behaviors.}

\textbf{\textit{The Nature of Non-reasoning Responses}:}
For proprietary reasoning models where only the final formulated answers are accessible, the episode allocation from the observable outputs is much closer to the non-reasoning group than to open-trace reasoning models. In other words, the externally visible responses exhibit limited allocation to exploration- and verification-associated episodes.

\textbf{\textit{Distillation Preserves Structure}:}
Within the distilled series, we observe that episode allocation profiles remain similar across model sizes. R1-Distill-Qwen-1.5B exhibits an episode distribution close to its teacher DeepSeek-R1 despite large differences in parameter count. This indicates that distillation can transfer not only answers but also episode-level reasoning structure, as reflected in allocation patterns.

\begin{table}[t]
    \centering
    \resizebox{\linewidth}{!}{
    \begin{tabular}{ccl|ccl}
    \toprule
    \multicolumn{3}{c|}{\textbf{ R vs. Reasoning (Answer)}} & \multicolumn{3}{c}{\textbf{ R vs. Non-Reasoning Models}} \\
    \midrule
    \textbf{Idx} & \textbf{Score} & \textbf{Pattern} & \textbf{Idx} & \textbf{Score} & \textbf{Pattern} \\
    \midrule
    1 & 0.2862 & Exp-Mon & 1 & 0.2510 & Exp-Mon \\
    2 & 0.2815 & Ver-Exp & 2 & 0.2405 & Mon-Exp \\
    3 & 0.2771 & Mon-Exp & 3 & 0.2073 & Exp-Ana-Imp \\
    4 & 0.2754 & Exp-Plan & 4 & 0.2033 & Ana-Exp-Ana \\
    5 & 0.2605 & Exp-Ver & 5 & 0.2028 & Exp-Ana-Exp \\
    6 & 0.2579 & Exp-Imp & 6 & 0.1974 & Exp-Ana \\
    7 & 0.2568 & Exp-Ana-Exp & 7 & 0.1917 & Ver-Exp \\
    8 & 0.2567 & Exp-Plan-Imp & 8 & 0.1838 & Exp-Ver \\
    9 & 0.2560 & Imp-Ver-Exp & 9 & 0.1830 & Exp-Mon-Exp \\
    10 & 0.2519 & Ana-Exp & 10 & 0.1806 & Ana-Exp-Mon \\
    \bottomrule
    \end{tabular}}
    \vspace{-2.2mm}
    \caption{
        Top discriminative episode $N$-grams ranked by Mutual Information (MI). Left: patterns that distinguish a reasoning model's reasoning trace from the final answer segment. Right: patterns that distinguish reasoning traces from non-reasoning model responses. \textbf{Higher MI indicates a stronger association between the pattern and the source group.}
    }
    \label{tab:ngram}
    \vspace{-4.2mm}
    \end{table}

\subsection{Transition Patterns}

Beyond marginal episode allocations, episode \emph{transitions} characterize how reasoning behaviors interact with each other. We convert each trace into a symbolic episode sequence (e.g., \textit{Read} $\rightarrow$ \textit{Plan} $\rightarrow$ \textit{Implement}) and use an information-theoretic criterion to identify distinctive local structures\footnote{Details in Appendix \ref{sec:implementation_details}}. Concretely, we compute the Mutual Information (MI) between the presence of episode $N$-grams and the source group, and extract the most discriminative patterns.
Formally, the MI between an episode $N$-gram pattern presence $X$ and the source group variable $G$ is defined as:
\begin{equation}
\small
I(X; G) = \sum_{x \in \{0,1\}} \sum_{g \in \mathcal{G}} 
p(x,g)\log \frac{p(x,g)}{p(x)\,p(g)},
\end{equation}
where $x \in \{0,1\}$ indicates whether a given episode $N$-gram is present in a trace, and $g$ denotes the source group. Since MI is symmetric and only measures the strength of association, we further use conditional probability to determine which group each top-$k$ discriminative pattern is attributed to.

\textit{\textbf{Reasoning vs.\ Answer Tokens:}}
Comparing the reasoning portion of a reasoning model to the final formulated answer tokens, we find that the most discriminative patterns are short feedback loops involving \textit{Explore}, \textit{Monitor}, and \textit{Verify} (Table~\ref{tab:ngram}). In particular, \textit{Exp-Mon} and \textit{Mon-Exp} rank among the top patterns, indicating frequent alternation between exploration and process monitoring during the reasoning trace. Patterns such as \textit{Ver-Exp} further suggest that verification is often followed by renewed exploration rather than immediate convergence. In contrast, the final answer tokens are characterized by more feed-forward transitions with limited recurrence of these evaluative loops.

\textit{\textbf{Reasoning vs.\ Non-Reasoning Models:}}
When comparing reasoning traces to responses from standard instruction-following models, we observe an overlapping set of discriminative patterns: non-reasoning responses are dominated by feed-forward transitions, where exploration-monitoring-verification loops are much less prevalent. This indicates that the gap between \textbf{reasoning and non-reasoning models is reflected not only in how much time is allocated to different behaviors, but also in the presence of recurrent transitions that interleave exploration with evaluation.}

\begin{table}[t]
    \centering
    \resizebox{0.48\textwidth}{!}{%
    \begin{tabular}{clc|clc}
        \toprule
        \multicolumn{3}{c|}{\textbf{Positive Correlation}} & \multicolumn{3}{c}{\textbf{Negative Correlation}} \\
        \midrule
        \textbf{\#} & \textbf{Feature} & \textbf{$\beta$} & \textbf{\#} & \textbf{Feature} & \textbf{$\beta$} \\
        \midrule
        1 & Trans (Exp $\to$ Mon) & +0.41 & 1 & Ratio (Exp) & -0.54 \\
        2 & Trans (Exp $\to$ Ana) & +0.31 & 2 & Trans (Exp $\to$ Ver) & -0.45 \\
        3 & Trans (Mon $\to$ Ana) & +0.28 & 3 & Trans (Exp $\to$ Ans) & -0.41 \\
        4 & Trans (Read $\to$ Ver) & +0.27 & 4 & Count (Total) & -0.36 \\
        5 & Ratio (Think) & +0.22 & 5 & Trans (Imp $\to$ Read) & -0.32 \\
        6 & Trans (Ans $\to$ Ver) & +0.22 & 6 & Trans (Imp $\to$ Imp) & -0.28 \\
        7 & Ratio (Ver) & +0.21 & 7 & Trans (Ana $\to$ Mon) & -0.26 \\
        8 & Trans (Read $\to$ Mon) & +0.18 & 8 & Trans (Ver $\to$ Read) & -0.25 \\
        9 & Trans (Read $\to$ Read) & +0.17 & 9 & Trans (Ver $\to$ Plan) & -0.25 \\
        10 & Ratio (Mon) & +0.16 & 10 & Trans (Mon $\to$ Read) & -0.22 \\
        \bottomrule
    \end{tabular}%
    }
    \vspace{-2.2mm}
    \caption{
        Top predictive episode-level features for correctness in a diagnostic Lasso logistic regression case study.
        Features are ranked by their coefficient magnitude ($\beta$), where the sign and magnitude of $\beta$ indicate the direction and relative strength of association with correctness under the fitted model.
    }
    \label{tab:correctness}
    \vspace{-4.2mm}
\end{table}

\begin{table*}[htb]
    \centering
    \resizebox{\textwidth}{!}{%
    \begin{tabular}{l | cccccccc | cccccccc | c}
    \toprule
    \textbf{Model} & \multicolumn{8}{c|}{\textbf{Percentage (\%)}} & \multicolumn{8}{c|}{\textbf{Token Count (\#)}} & \textbf{Avg} \\
    \cmidrule(lr){2-9} \cmidrule(lr){10-17}
     & \textbf{Read} & \textbf{Ana} & \textbf{Plan} & \textbf{Imp} & \textbf{Exp} & \textbf{Ver} & \textbf{Ans} & \textbf{Mon} & \textbf{Read} & \textbf{Ana} & \textbf{Plan} & \textbf{Imp} & \textbf{Exp} & \textbf{Ver} & \textbf{Ans} & \textbf{Mon} & \textbf{Tok} \\
    \midrule
    R1-Distill-Qwen-1.5B & \cellcolor{niceblue!39}4.18 & \cellcolor{niceblue!79}26.93 & \cellcolor{niceblue!39}4.20 & \cellcolor{niceblue!84}34.39 & \cellcolor{niceblue!63}13.23 & \cellcolor{niceblue!60}11.43 & \cellcolor{niceblue!37}3.80 & \cellcolor{niceblue!25}1.84 & \cellcolor{niceblue!71}414 & \cellcolor{niceblue!93}2666 & \cellcolor{niceblue!71}416 & \cellcolor{niceblue!96}3404 & \cellcolor{niceblue!85}1309 & \cellcolor{niceblue!83}1131 & \cellcolor{niceblue!69}376 & \cellcolor{niceblue!62}182 & 9897 \\
    \midrule
    L1 (1.5B) \cite{aggarwal2025l1controllinglongreasoning} & \cellcolor{niceblue!56}9.33 & \cellcolor{niceblue!69}18.34 & \cellcolor{niceblue!46}6.01 & \cellcolor{niceblue!84}34.03 & \cellcolor{niceblue!66}15.14 & \cellcolor{niceblue!50}6.99 & \cellcolor{niceblue!53}8.52 & \cellcolor{niceblue!23}1.64 & \cellcolor{niceblue!64}227 & \cellcolor{niceblue!72}447 & \cellcolor{niceblue!59}147 & \cellcolor{niceblue!80}829 & \cellcolor{niceblue!70}369 & \cellcolor{niceblue!61}170 & \cellcolor{niceblue!63}208 & \cellcolor{niceblue!44}40 & 2437 \\
    ThinkPrune (1.5B) \cite{hou2025thinkprunepruninglongchainofthought} & \cellcolor{niceblue!48}6.48 & \cellcolor{niceblue!72}20.75 & \cellcolor{niceblue!43}5.16 & \cellcolor{niceblue!88}39.39 & \cellcolor{niceblue!58}10.74 & \cellcolor{niceblue!53}8.37 & \cellcolor{niceblue!50}7.31 & \cellcolor{niceblue!24}1.80 & \cellcolor{niceblue!66}263 & \cellcolor{niceblue!80}841 & \cellcolor{niceblue!63}209 & \cellcolor{niceblue!87}1597 & \cellcolor{niceblue!72}436 & \cellcolor{niceblue!69}339 & \cellcolor{niceblue!67}297 & \cellcolor{niceblue!50}73 & 4054 \\
    A\&Z \citep{arora2025training} & \cellcolor{niceblue!43}5.12 & \cellcolor{niceblue!80}28.32 & \cellcolor{niceblue!41}4.69 & \cellcolor{niceblue!86}36.90 & \cellcolor{niceblue!54}8.76 & \cellcolor{niceblue!57}9.94 & \cellcolor{niceblue!41}4.63 & \cellcolor{niceblue!23}1.64 & \cellcolor{niceblue!66}272 & \cellcolor{niceblue!86}1505 & \cellcolor{niceblue!65}249 & \cellcolor{niceblue!89}1960 & \cellcolor{niceblue!73}465 & \cellcolor{niceblue!74}528 & \cellcolor{niceblue!65}246 & \cellcolor{niceblue!53}87 & 5312 \\
    \bottomrule
    \end{tabular}%
    }
    \vspace{-2.2mm}
    \caption{
        Cognitive behavior divergence of different efficient reasoning methods. L1 and ThinkPrune drastically reduce the budget for \textit{Verify} and \textit{Analyze}, while the last one maintains a distribution closer to the baseline.
    }
    \label{tab:efficient_ratio}
    \vspace{-4.2mm}
\end{table*}

\begin{table}[htb]
    \centering
    \setlength{\tabcolsep}{4pt}
    \resizebox{0.5\textwidth}{!}{
    \begin{tabular}{ccl|ccl|ccl}
    \toprule
    \multicolumn{3}{c|}{\textbf{R vs. L1}} & \multicolumn{3}{c|}{\textbf{R vs. ThinkPrune}} & \multicolumn{3}{c}{\textbf{R vs. A\&Z}} \\
    \cmidrule(lr){1-3} \cmidrule(lr){4-6} \cmidrule(lr){7-9}
    \textbf{Idx} & \textbf{Score} & \textbf{Pattern} & \textbf{Idx} & \textbf{Score} & \textbf{Pattern} & \textbf{Idx} & \textbf{Score} & \textbf{Pattern} \\
    \midrule
    1 & 0.3762 & N-V-N & 1 & 0.1506 & N-V-N & 1 & 0.1039 & M-E \\
    2 & 0.2491 & V-N-V & 2 & 0.1465 & V-N-I & 2 & 0.0928 & E-N-I \\
    3 & 0.2476 & M-N-V & 3 & 0.1240 & V-N & 3 & 0.0919 & M-E-I \\
    4 & 0.2291 & V-N & 4 & 0.1200 & I-N-V & 4 & 0.0911 & I-N-M \\
    5 & 0.2261 & M-V & 5 & 0.1120 & E-V-N & 5 & 0.0897 & N-M-I \\
    6 & 0.2233 & N-M & 6 & 0.1073 & V-N-V & 6 & 0.0836 & E-V-I \\
    7 & 0.2144 & I-V-M & 7 & 0.1070 & V-N-E & 7 & 0.0789 & V-A-M \\
    8 & 0.2075 & V-M & 8 & 0.1018 & I-V-N & 8 & 0.0746 & I-M-N \\
    9 & 0.2017 & I-N-M & 9 & 0.0924 & N-I-V & 9 & 0.0701 & E-I \\
    10 & 0.1949 & V-N-I & 10 & 0.0881 & V-M-P & 10 & 0.0692 & I-N-V \\
    \bottomrule
    \end{tabular}}
    \vspace{-2.2mm}
    \caption{
        Cognitive patterns that are lost in efficiency models compared to the baseline. L1 shows high distinctive scores, indicating a significant loss of complex verification loops (\textit{V-N-V}). Conversely, A\&Z shows much lower divergence scores, suggesting it preserves the baseline's topological structure more effectively.
    }
    \label{tab:efficient_ngram}
    \vspace{-4.2mm}
\end{table}

\section{Applications}

\subsection{How Episodes Correlate to Correctness}
\label{sec:episode_level_correctness}

To demonstrate the utility of our framework, we conduct a correctness-oriented case study examining how episode-level patterns are associated with solution correctness. Using a stratified sample of $500$ reasoning traces from the 5 representative open-source reasoning models, we formulate a binary classification setting that predicts answer correctness from quantitative cognitive features extracted from episode annotations\footnote{Details in Appendix \ref{sec:episode_level_diagnostics_of_correctness_details}}.

\paragraph{Methodology}
We map each reasoning trace to a feature vector consisting of three components: (i) global statistics (total token count, thinking-token count, and thinking-token ratio), (ii) episode intensities (token ratios for each of the eight episodes), and (iii) transition features (the flattened $8 \times 8$ episode transition matrix capturing frequencies of state shifts). 
Then, we fit a Lasso-regularized logistic regression model to predict the correctness of a reasoning trace:
\begin{equation}
\small
\mathcal{L} = -\sum_{i} \big[ y_i \log \hat{y}_i + (1 - y_i)\log(1 - \hat{y}_i) \big]
+ \lambda \|\mathbf{w}\|_1,
\end{equation}
where $\hat{y} = \sigma(\mathbf{w}^\top \mathbf{x})$, $\sigma(\cdot)$ denotes the sigmoid function and the $\ell_1$ penalty encourages sparsity in the coefficient vector. Table~\ref{tab:correctness} reports the top features ranked by coefficient magnitude ($\beta$). Here, each coefficient $\beta$ corresponds to a weight in $\mathbf{w}$ for a specific episode-level feature. A positive (negative) $\beta$ indicates that higher values of the feature are associated with increased (decreased) likelihood of a correct answer under this model, and the magnitude of $\beta$ reflects its relative importance among the selected features.

\paragraph{Results}
The most predictive positive features highlight how exploratory behavior is resolved during successful reasoning. In particular, \textit{Explore} $\rightarrow$ \textit{Monitor} and \textit{Explore} $\rightarrow$ \textit{Analyze} rank among the strongest positive coefficients, suggesting that correct solutions tend to route exploratory uncertainty into meta-level monitoring and renewed conceptual analysis. Similarly, \textit{Monitor} $\rightarrow$ \textit{Analyze} indicates that monitoring is often followed by additional analysis rather than immediate execution, consistent with an uncertainty-to-reasoning redirection pattern.
Verification-related transitions also appear as informative signals at different points in the trace. \textit{Read} $\rightarrow$ \textit{Verify}  and \textit{Answer} $\rightarrow$ \textit{Verify} suggest that traces associated with correctness more frequently include explicit checking both near the beginning and near the end.

On the negative side, a higher \textit{Explore} ratio is a strong risk indicator, suggesting that sustained exploration without subsequent stabilization is associated with incorrect outcomes. Two additional failure-associated patterns emerge. First, \textit{Explore} $\rightarrow$ \textit{Verify} reflects verification applied before exploration has been consolidated into a coherent hypothesis. Second, \textit{Implement} $\rightarrow$ \textit{Read} indicates breakdowns during execution that coincide with reverting to problem re-reading, a signal of disrupted reasoning flow.

This study illustrates \textbf{how episode-level representations can surface interpretable transition-level signatures associated with correctness}, complementing outcome-based evaluation with a structured diagnostic view of reasoning behavior. 

\paragraph{Further Discussion}
More importantly, these findings suggest that reasoning performance may be improved by explicitly shaping such structural patterns during training or inference. For instance, prior work has shown that encouraging cognitive behaviors such as backtracking and self-reflection can lead to more robust reasoning trajectories~\citep{kim2025astroteachinglanguagemodels, yang2025stepleapforwardselfbacktracking}. Similarly, incorporating key cognitive patterns into training data has been found to significantly enhance model performance~\citep{gandhi2025cognitivebehaviorsenableselfimproving}. Beyond training-time interventions, recent studies demonstrate that reasoning behaviors can also be steered at inference time via activation control, providing a practical pathway to translate diagnostic insights into performance gains~\citep{venhoff2025understandingreasoningthinkinglanguage}.
Taken together, these results highlight a promising direction: leveraging episode-level diagnostics not only for analysis, but also as a foundation for systematically improving reasoning processes.


\subsection{Episode Divergence for Efficient Models}

Efficient reasoning methods for LLMs are often evaluated primarily through surface metrics such as response length, leaving it unclear which reasoning behaviors are being altered. In this case study, we use \ours to characterize how different efficiency paradigms reshape episode-level dynamics. Specifically, we compare a baseline model (R1-Distill-Qwen-1.5B) against three representative strategies: (1) L1~\cite{aggarwal2025l1controllinglongreasoning}, (2) ThinkPrune~\cite{hou2025thinkprunepruninglongchainofthought}, and (3) the model proposed by~\citet{arora2025training}, referred as A\&Z.

Table~\ref{tab:efficient_ratio} summarizes how episode-level token allocation changes for these methods. Compared to the baseline, L1 and ThinkPrune substantially reduce the budget assigned to evaluative and conceptual episodes (e.g., \textit{Verify} and \textit{Analyze}) and shift the profile toward a more \textit{Implement}-heavy allocation. In contrast, A\&Z maintains an allocation profile closer to the baseline, retaining a substantial \textit{Verify} and \textit{Analyze} budget. These patterns suggest that efficiency methods are not uniformly compressive: they can induce qualitatively different redistributions of reasoning behaviors.

To further localize which transition-level structures are most affected, Table~\ref{tab:efficient_ngram} reports the most distinctive episode $N$-grams suppressed by each efficiency strategy relative to the baseline by computing the Mutual Information. L1 exhibits high divergence scores (peaking at 0.37), particularly for loop-like patterns such as \textit{N-V-N} (\textit{Analyze}$\rightarrow$\textit{Verify}$\rightarrow$\textit{Analyze}) and \textit{V-N-V}, indicating that recurrent verification loops are strongly attenuated. ThinkPrune shows a similar but generally weaker suppression of such loop structures. By contrast, A\&Z yields much lower divergence scores (peaking around 0.10), suggesting that it preserves more of the baseline transition topology while still reducing overall cost.

Overall, this case study illustrates that efficiency is not behaviorally neutral: \textbf{strategies that achieve similar surface-level reductions in length can correspond to markedly different changes in episode-level allocation and transition structure.}

\section{Results on Problems with Lower Difficulty Level}
\label{sec:gsm8k}

The analyses presented so far focus on Omni-Math, an olympiad-level benchmark that stresses reasoning at its limits. To examine whether the observed episode-level patterns are specific to this high-difficulty regime or reflect more general properties of model behavior, we apply \ours to a 50-problem subset of GSM8K~\citep{cobbe2021gsm8k}, a widely used benchmark of grade-school math word problems. We annotate four models: two reasoning models (DeepSeek-R1 and Qwen-3-32B (R)) and two non-reasoning models (Gemini-2.0-Flash and Qwen-3-32B). Table~\ref{tab:epi_ratio_gsm8k} reports the episode-level allocation. Several patterns emerge when comparing these results to the Omni-Math allocations in Table~\ref{tab:epi_ratio}.

\begin{table}[t]
    \centering
    \resizebox{\linewidth}{!}{%
    \begin{tabular}{l | cccccccc}
    \toprule
    \textbf{Model} & \textbf{Read} & \textbf{Ana} & \textbf{Plan} & \textbf{Imp} & \textbf{Exp} & \textbf{Ver} & \textbf{Ans} & \textbf{Mon} \\
    \midrule
    DeepSeek-R1       & \cellcolor{niceblue!53}21.82 & \cellcolor{niceblue!42}13.04 & \cellcolor{niceblue!30}4.13  & \cellcolor{niceblue!64}29.69 & \cellcolor{niceblue!34}6.88  & \cellcolor{niceblue!36}8.19  & \cellcolor{niceblue!44}14.88 & \cellcolor{niceblue!27}1.38 \\
    Qwen-3-32B (R)    & \cellcolor{niceblue!34}6.53  & \cellcolor{niceblue!32}5.70  & \cellcolor{niceblue!32}5.40  & \cellcolor{niceblue!46}16.48 & \cellcolor{niceblue!30}3.52  & \cellcolor{niceblue!98}55.89 & \cellcolor{niceblue!30}4.18  & \cellcolor{niceblue!28}2.30 \\
    \midrule
    Gemini-2.0-Flash  & \cellcolor{niceblue!44}14.81 & \cellcolor{niceblue!37}9.04  & \cellcolor{niceblue!30}3.85  & \cellcolor{niceblue!97}55.58 & \cellcolor{niceblue!5}0.00   & \cellcolor{niceblue!27}1.35  & \cellcolor{niceblue!44}14.62 & \cellcolor{niceblue!26}0.77 \\
    Qwen-3-32B        & \cellcolor{niceblue!54}21.94 & \cellcolor{niceblue!28}2.14  & \cellcolor{niceblue!55}22.91 & \cellcolor{niceblue!74}37.67 & \cellcolor{niceblue!5}0.00   & \cellcolor{niceblue!5}0.00   & \cellcolor{niceblue!42}13.40 & \cellcolor{niceblue!28}1.94 \\
    \bottomrule
    \end{tabular}%
    }
    \vspace{-2.2mm}
    \caption{
        Episode-level token allocation (\%) on a 50-problem subset of GSM8K.
        Rows are separated into reasoning (top) and non-reasoning (bottom) models.
        Compared to Omni-Math (Table~\ref{tab:epi_ratio}), all models shift more budget to \textit{Read} and \textit{Answer} on easier problems, while non-reasoning models largely suppress \textit{Explore} and \textit{Verify}.
    }
    \vspace{-4.2mm}
    \label{tab:epi_ratio_gsm8k}
\end{table}

\textbf{\textit{Easier problems shift budget toward reading and answering.}}
All four models allocate substantially more tokens to \textit{Read} on GSM8K than on Omni-Math, consistent with simpler problems requiring more time to process relative to the total response. \textit{Answer} proportions are also higher, reflecting shorter reasoning paths before commitment.

\textbf{\textit{Non-reasoning models largely suppress exploration and verification.}}
On Omni-Math, non-reasoning models already show limited \textit{Explore} and \textit{Verify} allocations; on GSM8K this suppression is near-total. Both Gemini-2.0-Flash and Qwen-3-32B assign essentially zero budget to \textit{Explore} and \textit{Verify}, suggesting that these models skip reflective checking entirely on problems they perceive as routine.

\textbf{\textit{Model-specific traits become more pronounced.}}
Cross-difficulty differences reveal characteristic model signatures. The reasoning mode of Qwen-3-32B allocates 55.89\% of its GSM8K budget to \textit{Verify}---far higher than on Omni-Math---suggesting it applies proportionally more checking effort on problems where execution is inexpensive. Gemini-2.0-Flash concentrates 55.58\% on \textit{Implement}, consistent with a direct-execution strategy that is amplified when problems offer little structural resistance.

Overall, these results show that the main qualitative distinctions between reasoning and non-reasoning models identified on Omni-Math are preserved, and in several respects sharpened, on medium-difficulty problems, supporting the generality of the episode-level framework across difficulty regimes.

\section{Conclusion}
In this work, we introduce \ours as an inductive, intermediate-scale framework that abstracts reasoning traces into functional reasoning steps grounded in cognitive theory. Through large-scale empirical analysis, we show that this abstraction makes previously opaque reasoning structures explicit, revealing consistent temporal organization and interpretable diagnostic patterns related to correctness and efficiency. Our framework provides a principled lens for analyzing, comparing, and diagnosing reasoning behavior in modern language models.

\section*{Limitations}
    Large-scale episode annotation relies on an automatic annotator, which may introduce labeling noise despite strong agreement with human annotations on a verified gold set. Moreover, our experiments focus primarily on mathematical problem solving; extending episode-level analysis to other domains and reasoning settings remains an important direction for future work.

\clearpage
\bibliography{custom}

\newpage
\appendix
\startcontents[appendix]
\printcontents[appendix]{ }{0}{\section*{Table of Contents for Appendix}}


\clearpage
\section{Related Work}
\label{sec:related_work}

\subsection{Cognitive Theories of Math Problem Solving}
Analyzing human problem-solving has evolved from broad cognitive taxonomies to domain-specific frameworks. Early models like Bloom's Taxonomy \citep{krathwohl2002revision} categorized cognitive processes hierarchically but failed to capture the iterative nature of mathematical problem-solving. Similarly, instructional frameworks such as \citet{polya1945how} four-phase model and subsequent refinements by \citet{mason2010thinking} and \citet{yeo2010characterising} provided valuable pedagogical scaffolding but lacked the fine-grained operationalization required for rigorous empirical annotation. Even more detailed models like \citet{greenes1995mathematics}, while emphasizing metacognition, remained too sequential to effectively code the nonlinear dynamics often observed in real-world reasoning tasks.

\citet{schoenfeld2014mathematical} episode theory offers a robust, empirically validated scheme for coding behaviors into distinct episodes such as \textit{Reading}, \textit{Analysis}, \textit{Exploration}, and \textit{Verification}. Unlike prescriptive models, Schoenfeld's framework explicitly captures the strategic decisions and metacognitive control—or lack thereof—that determine problem-solving success \citep{kuzle2013patterns}. This granular focus on cognitive transitions makes it particularly suitable for analyzing AI-generated reasoning, which often struggles with self-regulation. Consequently, Schoenfeld's model provides the necessary precision to systematically annotate and evaluate the complex, often non-linear reasoning traces investigated in this study. \citet{li-etal-2025-understanding} first applied this framework to analyze the reasoning process of large language models. However, they haven't conducted a systematic fine-grained analysis of the episode-level patterns of annotated reasoning traces or compared the episode-level patterns between different models.

\subsection{Large Reasoning Models}
The surge in LLM development has catalyzed substantial efforts to bolster their reasoning skills~\citep{ahn-etal-2024-large, besta2025reasoninglanguagemodelsblueprint, chen2025reasoningerasurveylong}. Most research has centered on enhancing these abilities via post-training methods. For instance, reinforcement learning has been utilized to steer models towards superior reasoning strategies~\citep{shao2024deepseekmathpushinglimitsmathematical, xiong2025selfrewardingcorrectionmathematicalreasoning, cui2025processreinforcementimplicitrewards,xiong2025deliberate,xiongenhancing}. Furthermore, instruction tuning using rigorously selected, high-quality datasets has proven effective in boosting performance~\citep{li-etal-2024-quantity, li-etal-2024-selective, ye2025limoreasoning, muennighoff2025s1simpletesttimescaling, li2025instruction，li2025instructiongrad}. Moreover,~\citet{snell2024scalingllmtesttimecompute, learningtoreason} have shown that scaling up test time compute can also improve the reasoning capabilities of models. Yet, while these models demonstrate remarkable gains in mathematical reasoning on benchmarks, there remains a notable gap in systematically understanding and quantifying how these enhancements alter model behavior. Recent efforts have begun to bridge this gap from an information-theoretic perspective; for example,~\citet{wang2025beyond} emphasize the critical role of high-entropy minority tokens in reasoning traces during reinforcement learning and Chain-of-Thought (CoT) paths. Distinct from their token-level, entropy-based methodology, our work analyzes model output behavior at the semantic and cognitive levels. Despite these divergent starting points, our findings are strikingly aligned: while their analysis reveals the importance of logical connectors within and across sentences, our study similarly highlights the positive contribution of "Monitor"-related phase transitions in the reasoning process.

\subsection{Efficient Reasoning Methods}
Overthinking~\cite{chen2025think23overthinkingo1like, fan2025missing} has been an identified issue of reasoning models. They tend to produce excessive intermediate steps or consider unnecessary details, which can lead to inefficient reasoning. To address this issue, several methods~\cite{sui2025stopoverthinkingsurveyefficient, feng2025efficientreasoningmodelssurvey} have been proposed to encourage the reasoning models to reason more efficiently. Specifically, L1~\cite{aggarwal2025l1controllinglongreasoning} proposes a simple reinforcement learning method that optimizes for accuracy and adherence to user-specified length constraints. ThinkPrune~\cite{hou2025thinkprunepruninglongchainofthought} offers a simple solution that continuously trains the long-thinking LLMs via reinforcement learning with an added token limit. \citet{arora2025training} train reasoning models to dynamically allocate inference-time compute based on task complexity. More works~\cite{fang2025thinklessllmlearnsthink, liu2025learnreasonefficientlyadaptive, xiang2025justthinkingefficientreasoning, li2025makes} have been proposed to encourage the efficient reasoning of models. However, currently, very limited work explicitly analyzes how these methods are different in behavior or episode-level patterns. Our work is the first to systematically analyze the episode-level patterns of these methods in comparison to the baseline reasoning models.

\subsection{Reasoning Analysis Methods}
Chain-of-thoughts (CoT)~\cite{wei2023chainofthoughtpromptingelicitsreasoning} can significantly elicit the reasoning capabilities of models. Various works have studied the different properties of CoT, including bias~\cite{wu2025doesreasoningintroducebias}, faithfulness~\cite{lanham2023measuringfaithfulnesschainofthoughtreasoning}, and redundancy~\cite{chen2025think23overthinkingo1like}. Specifically, for o1-like reasoning models that have particularly long reasoning traces and showcase more system-II level reasoning abilities, efforts have been made to uncover the structural patterns of CoT~\cite{jiang2025makesgoodreasoningchain}. \citet{bogdan2025thoughtanchorsllmreasoning} introduced a black-box method that measures each sentence's counterfactual importance to understand the sentence-level importance in long CoT chains. \citet{feng2025characterizeseffectivereasoningrevisiting}  introduced a graph view of CoT to extract structure and identified an effective statistic that correlated to the correctness of the reasoning trace.
\citet{li-etal-2025-understanding} and \citet{kargupta2025cognitivefoundationsreasoningmanifestation} incorporated theories from cognitive psychology to label the episodes of CoT in analogy of human problem-solving processes. 

Our work, built upon~\citet{li-etal-2025-understanding}, is the first to systematically analyze the statistical patterns of reasoning traces grounded in cognitive theories. While our work builds on a similar episode taxonomy, it introduces several key extensions. Rather than directly reusing the taxonomy in \citet{li-etal-2025-understanding}, we refine it to improve scalability and alignment with standardized LLM reasoning benchmarks by removing the hierarchical structure and introducing an explicit Answer state for practical applicability. In addition, whereas \citet{li-etal-2025-understanding} focuses on SAT-style data, we conduct our experiments on Omni-Math~\cite{gao2024omni}, a widely used dataset in contemporary LLM reasoning research, making our findings more representative of current evaluation settings. Beyond annotation, we perform a substantially deeper quantitative analysis of episode dynamics, including episode divergence, temporal evolution, coverage, and multi-step transition patterns. We further extend prior work by introducing episode-level n-gram pattern mining with mutual information, whereas \citet{li-etal-2025-understanding} primarily analyzes one-step transitions. Finally, we demonstrate broader downstream applications of the framework by using logistic regression to examine how episode-level features correlate with correctness, and by analyzing efficient reasoning methods to identify which behavioral patterns are selectively suppressed. Taken together, although both works are grounded in the same underlying cognitive theory, our approach represents a refined and extended framework with deeper analysis and stronger practical applicability.

Concurrent with our work, \citet{kargupta2025cognitivefoundationsreasoningmanifestation} also investigate cognitive behaviors in reasoning traces, with particular attention to success-related elements and temporal patterns. While both studies are grounded in cognitive science, they differ in scope and level of abstraction. \citet{kargupta2025cognitivefoundationsreasoningmanifestation} propose a taxonomy structured along four dimensions comprising 28 fine-grained cognitive elements, enabling detailed identification of localized reasoning behaviors. In contrast, our framework organizes reasoning into a smaller set of eight episodes that emphasize the overall structure and progression of problem solving. As such, their approach provides higher granularity at the micro level, whereas ours focuses on capturing the global dynamics of reasoning processes. We view these perspectives as complementary, collectively contributing to a more comprehensive understanding of cognitive behavior in large language model reasoning.

\clearpage
\section{Implementation Details}
\label{sec:implementation_details}

\subsection{Full Model List}
We conduct our analysis on a variety of reasoning and non-reasoning models. The reasoning models include DeepSeek-R1~\citep{deepseekai2025deepseekr1incentivizingreasoningcapability}, DeepSeek-R1-Distill-Qwen, QwQ-32B~\citep{qwq32b}, Phi4~\citep{abdin2024phi4}, Qwen3-32B~\citep{qwen3}, Gemini-2.5-Flash~\citep{comanici2025gemini25pushingfrontier}, GPT-o1-mini~\citep{openai2024o1mini} and GPT-o3-mini~\citep{openai2025o3mini}. The non-reasoning models include GPT-4o~\citep{openai2024gpt4ocard}, Gemini-2.0-Flash~\citep{geminiteam2024geminifamilyhighlycapable}, Qwen2.5-32B~\citep{qwen2.5} and the non-reasoning mode of Phi4 and Qwen3-32B. We also study some efficient reasoning models including L1~\citep{aggarwal2025l1}, ThinkPrune~\citep{hou2025thinkprunepruninglongchainofthought}, and models released by~\citet{arora2025training}.

\subsection{Temporal Dynamics Details}
To investigate the temporal dynamics of reasoning phases across model responses, we analyze how the distribution of cognitive phases evolves over the course of generation. For each annotated response, we divide the sequence into $B$ equal-sized temporal bins based on token position, where $B=25$ in our analysis. Within each bin, we compute the frequency of tokens belonging to each reasoning phase category.
Specifically, for a response with total length $L$ tokens, we define bin size as $\Delta = L / B$. Each token at position $t$ is assigned to bin $b = \lfloor t / \Delta \rfloor$, ensuring uniform temporal coverage across responses of varying lengths. For each category $c$ and bin $b$, we count the number of tokens belonging to that category, yielding a raw frequency distribution $f_{c,b}$.

To enable comparison across responses with different phase compositions, we normalize the frequency distribution within each response. For category $c$ in response $r$, we compute the normalized frequency in bin $b$ as:
\begin{equation}
\label{eq:normalized_freq}
    \tilde{f}_{c,b}^{(r)} = \frac{f_{c,b}^{(r)}}{\sum_{b'=1}^{B} f_{c,b'}^{(r)}}
\end{equation}
where the denominator represents the total number of tokens of category $c$ in response $r$. This normalization ensures that the distribution sums to 1 for each category within each response, allowing us to examine the relative temporal positioning of phases independent of their absolute frequency.

\subsection{Transition Patterns Details}
After annotation, each response can be represented with a sequence of cognitive phases. To study which phase combinations are specific to a certain model or kinds of models (e.g. Deepseek vs others, reasoning vs non-reasoning), we conduct the phase-based N-gram analysis that \textbf{treats each phase as a gram} and investigate what combinations of grams are most significant patterns of a type of models. Specifically, we use a letter to represent each phase (e.g. R for \textsc{Read}), and each response becomes a string. We use the mutual information to identify the most discriminative patterns between two groups:
\begin{equation}
\small
I(X; G) = \sum_{x \in \{0,1\}} \sum_{g \in \mathcal{G}} 
p(x,g)\log \frac{p(x,g)}{p(x)\,p(g)},
\end{equation}
where $x \in \{0,1\}$ indicates whether a given episode $N$-gram is present in a trace, and $g$ denotes the source group. Since MI is symmetric and only measures the strength of association, we further use conditional probability to determine which group each top-$k$ discriminative pattern is attributed to.

\clearpage
\section{Confusion Matrix}
\label{sec:confusion_matrix}

To further examine annotation quality, we report the confusion matrices of the four candidate annotation models on the same set of 7,067 human-annotated sentences in Tables~\ref{tab:confusion_matrix_gpt5}--\ref{tab:confusion_matrix_gemini_pro}. Rows correspond to gold labels and columns correspond to model predictions. Across all models, the matrices are strongly diagonally dominant, indicating that most sentences are assigned to the correct episode category.

\begin{table*}[h]
    \centering
    \resizebox{0.8\linewidth}{!}{%
    \begin{tabular}{lcccccccc}
        \toprule
        & \textbf{Analyze} & \textbf{Answer} & \textbf{Explore} & \textbf{Implement} & \textbf{Monitor} & \textbf{Plan} & \textbf{Read} & \textbf{Verify} \\
        \midrule
        \textbf{Analyze}   & 1031 & 14 & 35 & 225  & 7   & 10  & 18  & 67  \\
        \textbf{Answer}    & 1    & 405 & 2  & 4    & 2   & 1   & 0   & 1   \\
        \textbf{Explore}   & 6    & 2   & 594 & 13   & 10  & 14  & 0   & 15  \\
        \textbf{Implement} & 15   & 15  & 3  & 2542 & 0   & 4   & 1   & 86  \\
        \textbf{Monitor}   & 17   & 1   & 16 & 10   & 322 & 17  & 9   & 53  \\
        \textbf{Plan}      & 14   & 3   & 18 & 79   & 42  & 584 & 4   & 30  \\
        \textbf{Read}      & 6    & 0   & 2  & 14   & 1   & 6   & 259 & 1   \\
        \textbf{Verify}    & 25   & 18  & 7  & 89   & 8   & 1   & 4   & 996 \\
        \bottomrule
    \end{tabular}%
    }
    \caption{
        Confusion matrix of the GPT-5 annotation model on the 7,067 human-annotated sentences.
        Rows denote ground-truth labels and columns denote predicted labels.
    }
    \label{tab:confusion_matrix_gpt5}
    \vspace{-4.2mm}
\end{table*}

\begin{table*}[h]
    \centering
    \resizebox{0.8\linewidth}{!}{%
    \begin{tabular}{lcccccccc}
        \toprule
        & \textbf{Analyze} & \textbf{Answer} & \textbf{Explore} & \textbf{Implement} & \textbf{Monitor} & \textbf{Plan} & \textbf{Read} & \textbf{Verify} \\
        \midrule
        \textbf{Analyze}   & 1219 & 3   & 12  & 101  & 5   & 30  & 24  & 13  \\
        \textbf{Answer}    & 15   & 389 & 1   & 5    & 0   & 2   & 0   & 4   \\
        \textbf{Explore}   & 46   & 1   & 514 & 11   & 26  & 48  & 5   & 3   \\
        \textbf{Implement} & 102  & 9   & 5   & 2448 & 2   & 46  & 9   & 45  \\
        \textbf{Monitor}   & 5    & 1   & 6   & 2    & 409 & 14  & 2   & 6   \\
        \textbf{Plan}      & 18   & 0   & 7   & 42   & 13  & 680 & 12  & 2   \\
        \textbf{Read}      & 5    & 0   & 0   & 5    & 2   & 3   & 274 & 0   \\
        \textbf{Verify}    & 151  & 8   & 17  & 56   & 64  & 67  & 3   & 782 \\
        \bottomrule
    \end{tabular}%
    }
    \caption{
        Confusion matrix of the GPT-4.1 annotation model on the 7,067 human-annotated sentences.
    }
    \label{tab:confusion_matrix_gpt41}
    \vspace{-4.2mm}
\end{table*}

\begin{table*}[h]
    \centering
    \resizebox{0.8\linewidth}{!}{%
    \begin{tabular}{lcccccccc}
        \toprule
        & \textbf{Analyze} & \textbf{Answer} & \textbf{Explore} & \textbf{Implement} & \textbf{Monitor} & \textbf{Plan} & \textbf{Read} & \textbf{Verify} \\
        \midrule
        \textbf{Analyze}   & 1109 & 4   & 34  & 149  & 17  & 20  & 11  & 63  \\
        \textbf{Answer}    & 7    & 374 & 4   & 20   & 0   & 7   & 0   & 4   \\
        \textbf{Explore}   & 30   & 0   & 565 & 6    & 14  & 19  & 0   & 20  \\
        \textbf{Implement} & 68   & 1   & 8   & 2389 & 2   & 15  & 0   & 183 \\
        \textbf{Monitor}   & 17   & 3   & 31  & 9    & 299 & 22  & 2   & 62  \\
        \textbf{Plan}      & 22   & 4   & 36  & 62   & 13  & 593 & 5   & 39  \\
        \textbf{Read}      & 36   & 0   & 2   & 17   & 6   & 9   & 214 & 5   \\
        \textbf{Verify}    & 95   & 11  & 17  & 57   & 33  & 13  & 0   & 922 \\
        \bottomrule
    \end{tabular}%
    }
    \caption{
        Confusion matrix of the Gemini-2.5-Flash annotation model on the 7,067 human-annotated sentences.
    }
    \label{tab:confusion_matrix_gemini_flash}
    \vspace{-4.2mm}
\end{table*}

\begin{table*}[h]
    \centering
    \resizebox{0.8\linewidth}{!}{%
    \begin{tabular}{lcccccccc}
        \toprule
        & \textbf{Analyze} & \textbf{Answer} & \textbf{Explore} & \textbf{Implement} & \textbf{Monitor} & \textbf{Plan} & \textbf{Read} & \textbf{Verify} \\
        \midrule
        \textbf{Analyze}   & 1140 & 0   & 13  & 146  & 0   & 13  & 3   & 92  \\
        \textbf{Answer}    & 15   & 345 & 2   & 29   & 1   & 7   & 0   & 17  \\
        \textbf{Explore}   & 72   & 1   & 455 & 16   & 7   & 43  & 0   & 60  \\
        \textbf{Implement} & 94   & 0   & 3   & 2423 & 0   & 18  & 1   & 127 \\
        \textbf{Monitor}   & 22   & 2   & 13  & 14   & 229 & 31  & 1   & 133 \\
        \textbf{Plan}      & 69   & 1   & 10  & 55   & 7   & 580 & 5   & 47  \\
        \textbf{Read}      & 47   & 0   & 1   & 10   & 2   & 5   & 214 & 10  \\
        \textbf{Verify}    & 92   & 2   & 3   & 117  & 2   & 9   & 0   & 923 \\
        \bottomrule
    \end{tabular}%
    }
    \caption{
        Confusion matrix of the Gemini-2.5-Pro annotation model on the 7,067 human-annotated sentences.
    }
    \label{tab:confusion_matrix_gemini_pro}
    \vspace{-4.2mm}
\end{table*}

Several common patterns appear across all four annotators. First, the dominant confusions consistently involve behaviorally adjacent categories, especially \textit{Analyze} $\rightarrow$ \textit{Implement}, \textit{Verify} $\rightarrow$ \textit{Implement}, and, to a lesser extent, spillover from \textit{Verify} into other active reasoning categories such as \textit{Analyze}, \textit{Monitor}, and \textit{Plan}. Second, \textit{Answer} and \textit{Read} remain comparatively easy to identify, with much cleaner diagonals than the more process-oriented categories. Third, the stronger annotators mainly differ from the weaker ones not in the types of errors they make, but in how often they make them: GPT-5 and GPT-4.1 preserve the same overall structure while showing less off-diagonal mass than the two Gemini models. For GPT-5 specifically, the largest off-diagonal entries are \textit{Analyze} $\rightarrow$ \textit{Implement} (225 cases) and \textit{Verify} $\rightarrow$ \textit{Implement} (89 cases), which account for only 3.2\% and 1.3\% of the full dataset, respectively, or 4.4\% combined. Therefore, we believe the resulting annotations are sufficiently reliable for the downstream analyses in the main paper, including the temporal ``heartbeat'' patterns.

\clearpage
\section{Question-Domain Analysis}
\label{sec:question_domain_analysis}

To assess whether the ``heartbeat'' pattern in Figure~\ref{fig:temporal_plot} is driven only by aggregate averaging or also persists across different kinds of math problems, we analyze episode-level allocation separately for each major question domain in Omni-Math. Table~\ref{tab:domain_episode_allocation} reports the average proportion of tokens assigned to each episode category for each domain.

\begin{table*}[t]
    \centering
    \resizebox{\textwidth}{!}{%
    \begin{tabular}{lcccccccc}
        \toprule
        \textbf{Question Domain} & \textbf{Read} & \textbf{Analyze} & \textbf{Plan} & \textbf{Implement} & \textbf{Explore} & \textbf{Verify} & \textbf{Answer} & \textbf{Monitor} \\
        \midrule
        Algebra & 3.04 & 22.97 & 8.68 & 32.29 & 11.77 & 13.00 & 3.17 & 5.09 \\
        Calculus & 3.37 & 19.38 & 9.60 & 35.57 & 10.57 & 12.90 & 3.51 & 5.09 \\
        Applied Mathematics & 3.41 & 22.17 & 7.11 & 28.49 & 12.10 & 11.42 & 2.27 & 13.02 \\
        Discrete Mathematics & 3.26 & 23.99 & 7.46 & 28.32 & 11.67 & 14.01 & 2.79 & 8.49 \\
        Geometry & 3.18 & 21.56 & 8.23 & 30.42 & 9.69 & 19.58 & 2.26 & 5.09 \\
        Number Theory & 2.94 & 22.00 & 8.29 & 35.31 & 10.35 & 13.47 & 2.76 & 4.87 \\
        Precalculus & 2.90 & 20.62 & 9.51 & 27.99 & 11.82 & 16.30 & 5.56 & 5.32 \\
        \bottomrule
    \end{tabular}%
    }
    \vspace{-2.2mm}
    \caption{
        Average episode-level token allocation (\%) across question domains in Omni-Math.
        The domain-wise distributions are broadly consistent, while still revealing moderate variation in execution- and checking-related phases.
    }
    \label{tab:domain_episode_allocation}
    \vspace{-4.2mm}
\end{table*}

Several common patterns emerge across domains. First, \textit{Analyze} and \textit{Implement} consistently take the largest shares, indicating that most domains devote the bulk of reasoning to problem interpretation and execution. Second, \textit{Read} and \textit{Answer} remain comparatively small in every domain, suggesting that the main differences do not come from input parsing or final response formatting. Third, the largest domain-specific variation appears in the more execution- and control-oriented phases: \textit{Implement} is higher in Calculus and Number Theory, \textit{Verify} is especially prominent in Geometry and Precalculus, and \textit{Monitor} is notably higher in Applied Mathematics. Overall, these results suggest that the main ``heartbeat'' pattern is not an artifact of averaging across unrelated problem types; rather, it reflects a broadly shared reasoning structure with moderate domain-specific adjustments.

\clearpage
\section{Episode-level Diagnostics of Correctness Details}
\label{sec:episode_level_diagnostics_of_correctness_details}

In this section, we provide the detailed implementation of the correctness diagnostic analysis presented in Section~\ref{sec:episode_level_correctness}. We formulate the problem as a binary classification task, where we predict the correctness of a reasoning trace based on its episode-level cognitive features.

\subsection{Feature Engineering.} For each reasoning trace, we extract a comprehensive set of features capturing global statistics, episode intensity, and transition dynamics. Let $S = \{s_1, s_2, \dots, s_N\}$ be the sequence of sentences in a trace, where each sentence $s_i$ is associated with an episode tag $e_i \in \mathcal{E}$ and a token count $t_i$. The set of episode categories $\mathcal{E}$ consists of the 8 refined episodes: \textit{Read, Analyze, Plan, Implement, Explore, Verify, Monitor, Answer}. We compute the following feature groups:
\begin{itemize}
    \item \textbf{Global Statistics}:
    \begin{itemize}
        \item \textit{Total Tokens}: The total number of tokens in the response, $\sum t_i$, estimated using the GPT-4 tokenizer (`cl100k\_base`).
        \item \textit{Think Ratio}: The proportion of tokens belonging to the reasoning process (excluding the final \textit{Answer} content) relative to the total tokens.
    \end{itemize}
    \item \textbf{Episode Intensity (Token Ratios)}: For each episode category $c \in \mathcal{E}$, we compute the proportion of the total budget allocated to it:
    \[
    \text{Ratio}_c = \frac{\sum_{i: e_i = c} t_i}{\sum_{j} t_j}
    \]
    This yields 8 features representing the relative dominance of each cognitive behavior.
    \item \textbf{Transition Features}: We construct a transition matrix capturing the frequency of shifts between reasoning states. We compute the raw count of transitions from episode $src$ to episode $tgt$:
    \[
    \text{Trans}_{src \to tgt} = \sum_{i=1}^{N-1} \mathbb{I}(e_i = src \land e_{i+1} = tgt)
    \]
    This results in $8 \times 8 = 64$ transition features, capturing the structural flow of reasoning (e.g., \textit{Explore} $\to$ \textit{Verify}).
\end{itemize}

\subsection{Model Specification.} We employ a Lasso-regularized Logistic Regression model to identify the most predictive features while enforcing sparsity for interpretability. The model predicts the probability of correctness $p(y=1|\mathbf{x})$:
\[
p(y=1|\mathbf{x}) = \sigma(\mathbf{w}^\top \mathbf{x} + b)
\]
where $\mathbf{x}$ is the standardized feature vector, $\mathbf{w}$ are the learned coefficients, and $\sigma(\cdot)$ is the sigmoid function. To handle the high-dimensional feature space (particularly the transition matrix) and select only the most robust signals, we use L1 regularization (Lasso). The objective function is:
\begin{align*}
\min_{\mathbf{w}, b} \bigg( &-\frac{1}{M} \sum_{j=1}^M \Big[ y_j \log \hat{y}_j + (1 - y_j) \log (1 - \hat{y}_j) \Big] \\
&+ \lambda \|\mathbf{w}\|_1 \bigg)
\end{align*}
where $M$ is the number of samples and $\lambda$ controls the regularization strength.

\paragraph{Implementation Details.} We use the reasoning traces from 5 representative open-source reasoning models (DeepSeek-R1, DeepSeek-R1-Distill-Qwen-32B, Phi-4, Qwen3-32B, and QwQ-32B). All features are standardized using Z-score normalization before training. This ensures that the magnitude of coefficients directly reflects the relative importance of features. We use the \texttt{liblinear} solver which is well-suited for smaller datasets with L1 regularization. We set the regularization parameter $C=0.5$ (where $C = 1/\lambda$) to promote feature selection, and use a maximum of 2,000 iterations to ensure convergence. We analyze the learned coefficients $\mathbf{w}$. Features with positive coefficients increase the probability of a correct answer, while those with negative coefficients are associated with incorrect outcomes. Features with zero coefficients are pruned by the Lasso regularization, indicating they are less relevant for predicting correctness in this linear approximation.

\subsection{Full Feature Coefficients.} 
Table~\ref{tab:correctness_full} shows the full feature coefficients for the Lasso logistic regression model in the diagnostic Lasso logistic regression case study

\begin{table*}[ht]
    \centering
    \resizebox{0.7\textwidth}{!}{%
    \begin{tabular}{clc|clc}
        \toprule
        \multicolumn{3}{c|}{\textbf{Positive Contributors}} & 
        \multicolumn{3}{c}{\textbf{Negative Contributors}} \\
        \midrule
        \textbf{\#} & \textbf{Feature} & \textbf{$\beta$} & 
        \textbf{\#} & \textbf{Feature} & \textbf{$\beta$} \\
        \midrule
        1  & Trans (Exp $\to$ Mon)        & +0.41 & 1  & Ratio (Exp)                 & -0.54 \\
        2  & Trans (Exp $\to$ Ana)        & +0.31 & 2  & Trans (Exp $\to$ Ver)       & -0.45 \\
        3  & Trans (Mon $\to$ Ana)        & +0.28 & 3  & Trans (Exp $\to$ Ans)       & -0.41 \\
        4  & Trans (Read $\to$ Ver)       & +0.27 & 4  & Count (Total)               & -0.36 \\
        5  & Ratio (Think)                & +0.22 & 5  & Trans (Imp $\to$ Read)      & -0.33 \\
        6  & Trans (Ans $\to$ Ver)        & +0.22 & 6  & Trans (Imp $\to$ Imp)       & -0.28 \\
        7  & Ratio (Ver)                  & +0.21 & 7  & Trans (Ana $\to$ Mon)       & -0.26 \\
        8  & Trans (Read $\to$ Mon)       & +0.18 & 8  & Trans (Ver $\to$ Read)      & -0.25 \\
        9  & Trans (Read $\to$ Read)      & +0.17 & 9  & Trans (Ver $\to$ Plan)      & -0.25 \\
        10 & Ratio (Mon)                  & +0.16 & 10 & Trans (Mon $\to$ Read)      & -0.22 \\
        11 & Trans (Read $\to$ Imp)       & +0.15 & 11 & Ratio (Ans)                 & -0.21 \\
        12 & Trans (Exp $\to$ Read)       & +0.14 & 12 & Trans (Plan $\to$ Mon)      & -0.19 \\
        13 & Trans (Plan $\to$ Ver)       & +0.12 & 13 & Trans (Read $\to$ Ans)      & -0.18 \\
        14 & Trans (Plan $\to$ Plan)      & +0.12 & 14 & Trans (Imp $\to$ Ans)       & -0.15 \\
        15 & Ratio (Plan)                 & +0.10 & 15 & Trans (Ana $\to$ Ana)       & -0.14 \\
        16 & Trans (Imp $\to$ Mon)        & +0.07 & 16 & Ratio (Read)                & -0.13 \\
        17 & Trans (Ana $\to$ Ver)        & +0.07 & 17 & Trans (Ans $\to$ Ana)       & -0.11 \\
        18 & Trans (Ver $\to$ Imp)        & +0.05 & 18 & Trans (Read $\to$ Exp)      & -0.11 \\
        19 & Trans (Mon $\to$ Ver)        & +0.05 & 19 & Trans (Ver $\to$ Mon)       & -0.08 \\
        20 & Trans (Ana $\to$ Plan)       & +0.04 & 20 & Trans (Exp $\to$ Plan)      & -0.08 \\
        21 & Trans (Mon $\to$ Mon)        & +0.04 & 21 & Trans (Plan $\to$ Imp)      & -0.06 \\
        22 & Trans (Plan $\to$ Read)      & +0.03 & 22 & Trans (Ans $\to$ Imp)       & -0.05 \\
        23 & Trans (Plan $\to$ Exp)       & +0.01 & 23 & Trans (Ans $\to$ Exp)       & -0.05 \\
           &                               &       & 24 & Trans (Imp $\to$ Ana)       & -0.03 \\
           &                               &       & 25 & Trans (Read $\to$ Plan)     & -0.03 \\
           &                               &       & 26 & Trans (Ans $\to$ Mon)       & -0.02 \\
           &                               &       & 27 & Trans (Ver $\to$ Ana)       & -0.01 \\
        \bottomrule
    \end{tabular}%
    }
    \caption{
        The full predictive episode-level features for correctness in the diagnostic Lasso logistic regression case study.
        Features are ranked by their coefficient ($\beta$) magnitude.
    }
    \label{tab:correctness_full}
\end{table*}

\clearpage
\section{Detailed \ours Framework}
\label{sec:detailed_ours_framework}

\begin{figure*}[h]
    \centering
    \includegraphics[width=0.6\linewidth]{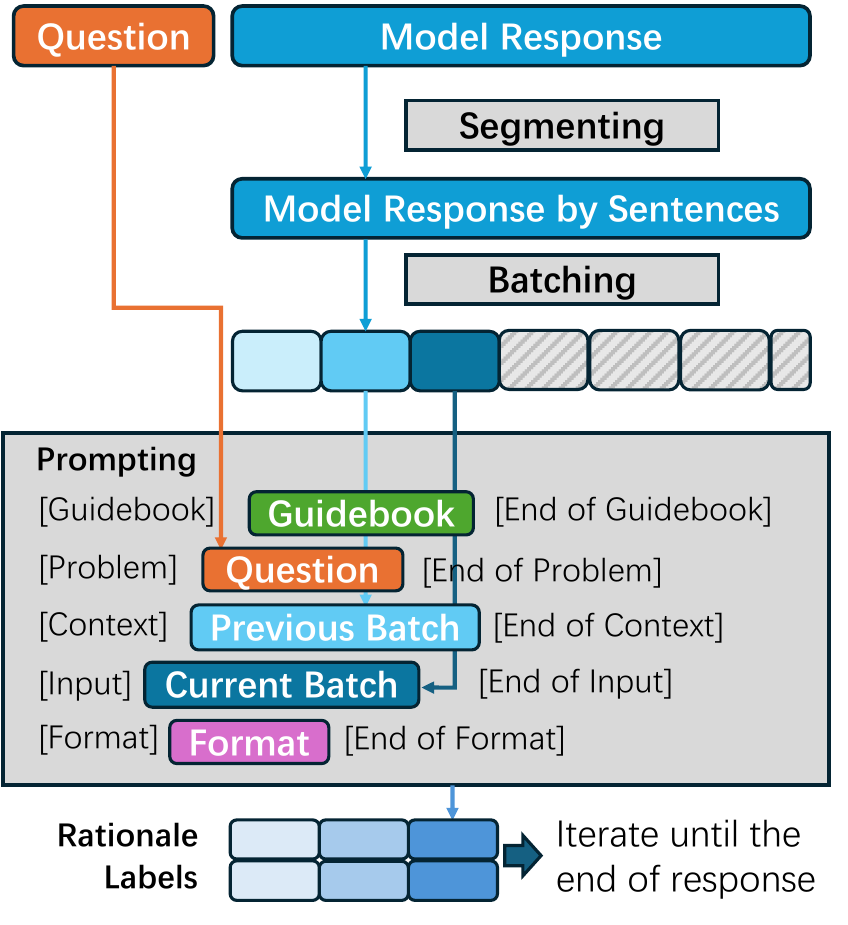}
    \caption{The \ours Framework. For each question-response pair, the model response is first segmented into sentences. They are then tagged by the annotation models in batches, along with information about the guidebook, question, context, and format. The guidebook is in Appendix~\ref{sec:refined_annotation_guidebook}, the prompt template is in Appendix~\ref{sec:annotation_prompt}.}
    \label{fig:pipeline}
\end{figure*}

The detailed \ours Framework is shown in Figure~\ref{fig:pipeline}. For each batch of sentences in the model response, the annotation model tags the episode category for each sentence based on the guidebook, question, previous context and outputs the rationale and annotation in JSON format as instructed by the format prompt. The labels after batch processing are then concatenated to form the labels for the whole response. The guidebook and prompt template in the figure are in Appendix~\ref{sec:refined_annotation_guidebook} and Appendix~\ref{sec:annotation_prompt}.

\clearpage
\section{Refined Annotation Guidebook}
\label{sec:refined_annotation_guidebook}

In this project, we aim to analyze the reasoning process of current large language models (LLMs) with advanced reasoning capabilities, i.e., Large Reasoning Models (LRMs), based on a modified version of Alan Schoenfeld's (1985) ``Episode-Timeline'' framework for problem-solving. The original Schoenfeld theory was built on hundreds of hours of recorded tapes of students tackling non-routine math problems while being asked to think aloud. Widely regarded as a gold-standard framework in mathematics education research, this theory offers a rigorously validated, fine-grained lens for dissecting both expert and novice problem-solving strategies. After thorough investigation, we find that the thinking process of LRMs can be well-aligned with the episodes in the theory, as they also follow similar problem-solving processes. Thus, in this project, we aim to annotate the model (solver) responses with these episode categories. To better apply the theory to the analysis of model responses, we utilize sentence-level annotation, which is used to capture the fine-grained behavior of each sentence, including \textbf{eight} categories: Read, Analyze, Plan, Implement, Explore, Verify, Monitor, and Answer. The original Schoenfeld theory only has six categories: Read, Analyze, Plan, Implement, Explore, and Verify. These categories describe the thinking behaviors of how humans solve a problem. Later, an additional ``Monitor'' category was included in the system to capture behaviors that do not contain specific content but are still important, such as ``Let me think.'' Moreover, when trying to apply the theory to analyzing LLM behaviors, we introduce another category, ``Answer,'' to represent the sentence that delivers the answer. Thus, in total, there are eight categories. For each sentence, the annotation depends on both the current sentence itself and its context.

\subsection{Label Definitions and Guidelines}

\paragraph{1. Read}
\begin{itemize}
    \item \textbf{Definition:} This is usually the initial phase, which focuses on extracting or restating the given information, conditions, and the goal of the problem as presented. It involves understanding the question without any inference of strategy or reasoning.
    \item \textbf{Guidelines:}
    \begin{itemize}
        \item Sentences in this category should directly present the content of the original problem statement.
        \item Look for phrases that recall or repeat elements of the question.
        \item This label is mostly presented for the model's initial processing of the problem.
    \end{itemize}
    \item \textbf{Potential Keywords/Indicators:} ``The question asks...'', ``The problem requires...'', ``We are given...'', ``The goal is to...'', ``The choices are...'', direct quotes from the problem.
    \item \textbf{Distinguishing Features:}
    \begin{itemize}
        \item This stage is purely about understanding the input, not about processing it or deciding how to solve it. It should not contain content like trying to understand or analyze the question.
        \item Avoid labeling sentences as Read if they include any form of analysis or evaluation of the problem. The Read stage usually appears at the beginning of the reasoning. However, it can also appear in the middle of the reasoning, in order to ensure that the question was understood correctly.
    \end{itemize}
    \item \textbf{Example:} ``The question asks us to find the value of x in the equation $2x + 5 = 10$.''
\end{itemize}

\paragraph{2. Analyze}
\begin{itemize}
    \item \textbf{Definition:} This stage involves constructing or recalling relevant theories, introducing necessary symbols, and deducing relationships based on the problem statement and existing knowledge. The core activity is explanation or logical inference that sets the stage for the solution but does not involve concrete calculations yet.
    \item \textbf{Guidelines:}
    \begin{itemize}
        \item Sentences should explain the underlying mathematical concepts or principles relevant to the problem.
        \item This category includes analysis of either the problem itself or intermediate results.
        \item This label applies to logical deductions and inferences made with certainty.
    \end{itemize}
    \item \textbf{Potential Keywords/Indicators:} ``According to...'', ``We can define...'', ``This implies that...'', ``Therefore...'', ``Based on this...'', ``We can infer that...'', ``Let's note that...'', ``Let me observe that...'', ``Let's recall that...''
    \item \textbf{Distinguishing Features:}
    \begin{itemize}
        \item The Analyze episode involves certain inferences and explanations, unlike Explore, which shows uncertainty.
        \item The actual execution of calculations is mainly in the Implement stage.
        \item Analyze does not involve any concrete calculation, which is unlike Implement.
        \item The behavior of setting some basic notations should be in the Implement stage, e.g., ``Let me set $a=1$, then...''.
    \end{itemize}
    \item \textbf{Important Note:} Be careful not to include sentences that involve substituting values or performing calculations, as those belong to the Implement stage.
    \item \textbf{Example:} ``According to the Pythagorean theorem, in a right-angled triangle, the square of the hypotenuse is equal to the sum of the squares of the other two sides.'' or ``If I can get the equation in slope-intercept form ($y = mx + b$), then I can plug in $y = 4$ and solve for $x$, which should be $d$.''
\end{itemize}

\paragraph{3. Plan}
\begin{itemize}
    \item \textbf{Definition:} This stage involves announcing the next step or outlining the entire solution strategy. It represents a commitment to a particular course of action before the actual execution begins.
    \item \textbf{Guidelines:}
    \begin{itemize}
        \item Sentences should clearly state the intended next step or the overall plan.
        \item Look for explicit declarations of intent, often using the first person or imperative voice.
        \item This stage signifies that a decision has been made on how to proceed, and the next step should be related to math problem solving, rather than generally saying ``let's think about it.''
    \end{itemize}
    \item \textbf{Potential Keywords/Indicators:} ``Next, we will...'', ``The next step is to...'', ``We need to...'', ``Let's proceed by...'', ``I will now...'', ``The plan is to...'', ``We should first...'', ``To..., do...'', ``The xxx we need/want is...'', ``Let's...'', ``Then/Now calculate/consider...''.
    \item \textbf{Distinguishing Features:}
    \begin{itemize}
        \item The Plan phase clearly indicates the intended action, unlike Analyze, which explains concepts, or Explore, which suggests possibilities.
        \item It precedes the actual carrying out of the plan in the Implement stage.
        \item Note that sentences like ``Let's denote...'' are Analyze, because this is introducing a new variable, rather than making a plan.
        \item Sentences like ``let's verify...'', ``let's test...'' or ``let's double-check'' are Verify.
    \end{itemize}
    \item \textbf{Example:} ``Next, we will differentiate both sides of the equation with respect to $x$.''
\end{itemize}

\paragraph{4. Implement}
\begin{itemize}
    \item \textbf{Definition:} This stage is the operational phase where the planned strategy is executed. It involves setting up basic notations, performing specific calculations, constructing diagrams, enumerating possibilities, or coding solutions using numerical values, symbols, or geometric objects.
    \item \textbf{Guidelines:}
    \begin{itemize}
        \item Sentences should describe the actual steps taken to solve the problem.
        \item Look for mathematical operations, substitutions, and the generation of intermediate results.
        \item This stage is about ``doing'' the math.
    \end{itemize}
    \item \textbf{Potential Keywords/Indicators:} ``Substituting $x = 2$, we get...'', ``Therefore, $P(1) = -1$'', ``Expanding the expression...'', ``The matrix becomes...'', ``Let me set $a=1$, then...'', ``Let's denote $x=10$...'', actual mathematical equations and calculations.
    \item \textbf{Distinguishing Features:}
    \begin{itemize}
        \item Implement involves concrete actions and calculations, unlike Analyze, which focuses on theoretical explanations, or Plan, which outlines future actions.
        \item If a conclusion follows the implementation of math, that conclusion is tagged as Implement, such as ``therefore, the sum of all possible values is 5.''
    \end{itemize}
    \item \textbf{Example:} ``Substituting $x = 3$ into the equation, we get $2(3) + 5 = 6 + 5 = 11$.''
\end{itemize}

\paragraph{5. Explore}
\begin{itemize}
    \item \textbf{Definition:} This stage is characterized by generating potential ideas, making guesses, drawing analogies, or attempting trial calculations that might be abandoned later. The model is exploring different avenues without committing to a specific solution path. This stage often involves uncertainty.
    \item \textbf{Guidelines:}
    \begin{itemize}
        \item Sentences should suggest alternative approaches or possibilities.
        \item Look for tentative language and expressions of uncertainty.
        \item This stage involves brainstorming and initial investigations without a clear commitment to a particular method.
    \end{itemize}
    \item \textbf{Potential Keywords/Indicators:} ``Maybe we can try...'', ``Perhaps we could use...'', ``What if we consider...'', ``Another possibility is...'', ``Could this be related to...'', ``Maybe I should...'', ``Maybe there is another way...'', ``Maybe we can try...'', ``Maybe there is a better way...'', ``Maybe consider...'', ``Perhaps... is...'', ``Let's try...'', ``Alternatively, maybe use...'', ``Wait, but maybe...'', ``But in soccer, it's possible to lose a game but still have more total goals?''; question marks indicating uncertainty about a step.
    \item \textbf{Distinguishing Features:}
    \begin{itemize}
        \item Explore is marked by uncertainty and a lack of commitment, unlike Plan, which announces a definite course of action.
        \item It involves considering various options before settling on a specific plan. If a sentence contains analyzing the problem, implementing the calculation, or verifying the result or thought, even if it follows sentences like ``Maybe we can try...'', the sentences are not considered Explore at the sentence level, and therefore should not be labeled as Explore. Rather, these sentences are considered Analyze, Implement, or Verify within the Explore episode at the paragraph level. Only sentences like ``Maybe we can try...'' will be labeled as Explore at the sentence level.
    \end{itemize}
    \item \textbf{Example:} ``Maybe we can try substituting different values for $x$ to see if we can find a pattern.''
\end{itemize}

\paragraph{6. Verify}
\begin{itemize}
    \item \textbf{Definition:} This stage involves judging the correctness, effectiveness, or simplicity of the obtained result or the method used. It might include checking the answer, using an alternative method for calculation, or estimating bounds.
    \item \textbf{Guidelines:}
    \begin{itemize}
        \item Sentences should express an evaluation or confirmation of the solution or the process.
        \item Look for keywords related to checking, confirming, or validating.
        \item This stage ensures the solution and result are accurate and make sense.
    \end{itemize}
    \item \textbf{Potential Keywords/Indicators:} ``Let me double-check...'', ``This is consistent with...'', ``Plugging it back in...'', ``Therefore, the answer is correct.'', ``Let's confirm...'', ``Let me check again...'', ``We can confirm this by...'', ``This result seems reasonable because...'', ``The answer is...?'', ``Is the answer...?'', ``Is there any mistake?'', ``Did I make a mistake?'', ``This is the same/correlated as previous...'', ``But this seems to contradict...'', ``...lead/arrive to the same answer'', ``Wait, we don't know... yet'', ``Let's try another way to verify...'', ``XXX is possible/impossible.'' When the following sentences are meant as conclusions, ``...is indeed...'', ``...should be...''
    \item \textbf{Distinguishing Features:}
    \begin{itemize}
        \item Verify focuses on evaluating the solution, unlike Implement, which focuses on generating it.
        \item It often involves comparing the result with initial conditions or using alternative methods.
    \end{itemize}
    \item \textbf{Example:} ``Let me double-check my calculations: $2 \times 3 + 5 = 11$, which matches the previous result.''
\end{itemize}

\paragraph{7. Monitor}
\begin{itemize}
    \item \textbf{Definition:} This additional category captures sentences that are typically short interjections or expressions indicating the model's self-monitoring, hesitation, or reflection at the juncture between different episodes. These often do not contain substantial problem-solving content and are brief pauses in the thought process.
    \item \textbf{Guidelines:}
    \begin{itemize}
        \item Sentences should be short phrases indicating a shift in thought or a brief pause.
        \item Look for expressions of uncertainty, reflection, or transition.
        \item This label is for meta-comments that don't fit neatly into the other problem-solving stages.
    \end{itemize}
    \item \textbf{Potential Keywords/Indicators:} ``Hmm...'', ``Wait...'', ``Let me think.'', ``Okay...'', ``Let's see.'', ``Hold on.'', ``But wait, hold on.''
    \item \textbf{Distinguishing Features:}
    \begin{itemize}
        \item Monitor sentences lack the substantive content of the other categories and primarily serve as indicators of the model's internal processing flow.
        \item They are often very short and act as bridges between more content-heavy stages.
        \item In most cases, it should not contain evaluation for previous steps or contain any specific question or solution content. If it contains specific content, e.g., ``Wait, the problem says...'', it should be categorized into others like Read.
    \end{itemize}
    \item \textbf{Example:} ``Wait.''
\end{itemize}

\paragraph{8. Answer}
\begin{itemize}
    \item \textbf{Definition:} This stage is used for sentences that explicitly state an answer or conclusion to the problem. These sentences deliver the result, either as a final answer at the end of the response or as an intermediate answer that may be subject to later verification or revision. Note: it should be the answer to the given problem, rather than an intermediate answer for a calculation step.
    \item \textbf{Guidelines:}
    \begin{itemize}
        \item Sentences should directly present a solution, value, or conclusion in response to the given problem statement.
        \item Look for clear, declarative statements that summarize the outcome of the reasoning or calculation.
        \item This category applies whether the answer is final or provisional.
    \end{itemize}
    \item \textbf{Potential Keywords/Indicators:} ``The answer is...'', ``Hence, the result is...'', ``So, the final answer is...''.
    \item \textbf{Distinguishing Features:}
    \begin{itemize}
        \item Answer sentences are characterized by their directness in providing a result to the given problem, unlike Verify, which focuses on checking correctness, or Implement, which details the process of obtaining the result.
        \item These sentences often appear at the end of a solution but can also occur mid-response as provisional answers.
    \end{itemize}
    \item \textbf{Example:} ``Therefore, the answer is 24.''
\end{itemize}

\subsection{Important Considerations for Annotators}
\begin{itemize}
    \item \textbf{Sentence-Level Focus:} Annotate each sentence individually based on its primary function within the problem-solving process.
    \item \textbf{Context is Key:} While keywords can be helpful, always consider the context of the sentence within the overall response. A sentence might contain a keyword but function differently based on the surrounding text.
    \item \textbf{Refer to Examples:} The examples provided in this guidebook and any additional examples you encounter should serve as valuable references.
\end{itemize}

\clearpage
\section{Annotation Prompt}
\label{sec:annotation_prompt}
We use the prompt template in Figure~\ref{fig:prompt} to annotate the reasoning traces in our study. The detailed content of the guidebook can be found in Appendix~\ref{sec:refined_annotation_guidebook}.

\begin{figure*}[h]
\begin{tcolorbox}[
  enhanced, 
  colframe=cyan!75!black, 
  colback=white, 
  coltitle=white, 
  colbacktitle=cyan!75!black, 
  width=\linewidth, 
  arc=2mm, 
  auto outer arc, 
  boxrule=0.5pt, 
  left=10pt, 
  right=10pt, 
  drop shadow={black!50!white},
  top=10pt, 
  bottom=10pt, 
  title=\textbf{Prompt Template for Annotation}, 
  fonttitle=\bfseries, 
  title code={ ode[rounded corners, fill=blue!75!black, draw=none, text=white] at (frame.title) {\textbf{xxx}};}, 
  attach boxed title to top center={yshift=-2mm}, 
  boxed title style={sharp corners, size=small}, 
]
In this project, we aim to analyze the reasoning process of current large language models (LLMs) with advanced reasoning capabilities, i.e., Large Reasoning Models, LRMs, based on a modified version of Alan Schoenfeld's (1985) "Episode-Timeline" framework for problem-solving. Given the model response you need to annotate the sentence-level behavior of the model response with the eight categories: Read, Analyze, Explore, Plan, Implement, Verify, Monitor, and Answer.
\\

The [Guidebook] - [End of the Guidebook] section provides the detailed introduction and definition of each category. The [Math Problem] - [End of the Math Problem] section provides a math problem.

The [Overall Response] - [End of the Overall Response] section provides the overall response of the model to the math problem.

The [Previous Context] - [End of the Previous Context] section provides all the previous context of the response that has been annotated and their corresponding labels.

The [Input] - [End of the Input] section provides the sentences that need to be annotated.

The [Format] - [End of the Format] section provides the format of the output.
\\

[Guidebook]

\textbf{(Guidebook Content)}

[End of Guidebook]

[Math Problem]

\textbf{(The Question)}

[End of the Math Problem]

[Previous Context]

The previous sentences are:\textbf{(Previous batch sentences)}

[End of Previous Context]

[Input]

The following sentences that you need to classify:\textbf{([1] xxx [2] xxx ...  [batch\_num] xxx)}

[End of Input]

[Format]

You should format the output in JSON format regarding the index, a short rationale and the fine-grained class of the indexed sentence. 
The format is as follows:" \\
\{'sentences': [\\
    \{'index'\: 'The index of the sentence', 'reason': 'The short reason of the classification', 'category': 'The fine-grained class of the sentence'\},\\
    \{'index': 'The index of the sentence', 'reason': 'The short reason of the classification', 'category': 'The fine-grained class of the sentence'\},\\
    ...\\
  ]\}" 

[End of Format] \\

Now, annotate the sentences in the [Input] - [End of the Input] section. Refer to the guidebook to make the decision. Strictly follow the index number of the sentence in the [Input] - [End of the Input] section for labeling. You should output the label for batch\_num sentences.

\end{tcolorbox}
\caption{The prompt template we used to annotate the reasoning episode.}

\label{fig:prompt}
\end{figure*}

\end{document}